\documentclass{article}

    \PassOptionsToPackage{numbers, compress}{natbib}
 \usepackage[eandd, final]{neurips_2026}   % CAMERA-READY (non-anonymous)
\usepackage[utf8]{inputenc} % allow utf-8 input
\usepackage[T1]{fontenc}    % use 8-bit T1 fonts
\usepackage{hyperref}       % hyperlinks
\usepackage{url}            % simple URL typesetting
\usepackage{booktabs}       % professional-quality tables
\usepackage{amsfonts}       % blackboard math symbols
\usepackage{nicefrac}       % compact symbols for 1/2, etc.
\usepackage{microtype}      % microtypography
\usepackage{xcolor}         % colors

\usepackage{graphicx}
\usepackage{booktabs}
\usepackage{microtype}
\usepackage{multirow}
\usepackage{array}
\usepackage{booktabs}
\usepackage{amsmath}
\usepackage{tabularx}
\usepackage[table]{xcolor}
\usepackage{adjustbox}
\usepackage{tabularx}
\usepackage{wrapfig}
\usepackage{placeins}
\usepackage{subcaption}

\usepackage{enumitem}
\setlist[itemize]{leftmargin=*}

\newcolumntype{L}[1]{>{\raggedright\arraybackslash}m{#1}}
\newcolumntype{R}[1]{>{\raggedleft\arraybackslash}m{#1}}

\title{AgroOmni: A Large-Scale Multi-view Agricultural Dataset for Cross-Scale Multimodal Reasoning}

\author{%
  Jiarui Zhang$^{1,*}$ \quad
  Junqi Hu$^{1,*}$ \quad Zurong Mai$^{1,*}$ \quad Yang Liu$^{2,\dagger}$ \quad Yuhang Chen$^{1}$ \\
  \textbf{Shuohong Lou$^{1}$ \quad Henglian Huang$^{1}$ \quad Hong Cheng$^{2}$ \quad Lingyuan Zhao$^{3}$ \quad Jianxi Huang$^{4,5,6}$} \\
  \textbf{Yutong Lu$^{1,7}$ \quad Haohuan Fu$^{7,8}$ \quad Juepeng Zheng$^{1,7,\dagger}$} \\
  $^{1}$Sun Yat-sen University \quad
  $^{2}$The Chinese University of Hong Kong \\
  $^{3}$HuanTian Wisdom Technology Co., Ltd. \quad
  $^{4}$China Agricultural University \\
  $^{5}$Southwest Jiaotong University \quad
  $^{6}$Guangxi University of Science and Technology \\
  $^{7}$National Supercomputing Center in Shenzhen \quad
  $^{8}$Tsinghua Shenzhen International Graduate School
}

\begin{document}

\maketitle

% Unnumbered footnote for the author marks (no marker of its own).
\begingroup
\renewcommand{\thefootnote}{}
\footnotetext{$^{*}$ Equal contribution.\quad $^{\dagger}$ Corresponding authors.}
\endgroup

\begin{abstract}
    Modern agricultural data is sourced from diverse platforms and spans multiple spatial scales, ranging from ground-level close-up photography to Unmanned Aerial Vehicle (UAV) aerial observation and satellite remote sensing imagery. Accordingly, agricultural multimodal reasoning demands robust cross-scale spatial understanding. However, due to the lack of multi-view agricultural benchmark datasets, existing multimodal large language models (MLLMs) exhibit severe ground-level bias, which leads to scale confusion then semantic collapse in agricultural perception tasks, such as misinterpreting farmland imagery as walls or floors. To address this, we introduce  \textbf{AgroOmni}, a large-scale multi-view training corpus with 268,119 Visual Question Answering pairs covering 56 specialized task categories across 14 task types, designed to capture diverse scales in modern precision agriculture.  % Built on this dataset, we propose \textbf{AgroNVILA}, an MLLM that utilizes the Perception-Reasoning Decoupling ({PRD}) architecture. On the perception side, we incorporate a View-Conditioned Meta-Net ({VCMN}), which injects macroscopic spatial context into visual tokens, resolving scale ambiguities with minimal computational overhead. On the reasoning side, Agriculture-aware Relative Policy Optimization ({ARPO}) leverages reinforcement learning to align the model's decision-making with expert agricultural logic, preventing statistical shortcuts. 
    Built on this dataset, we propose AgroNVILA, which achieves a new state-of-the-art of 59.86\% on the AgroMind benchmark ($+12.57\%$ over GPT-5.2), effectively mitigating the multi-view cross-scale gap for holistic agricultural understanding. Diagnostic evaluations on AgMMU further reveal an inherent heterogeneity between macro-priors and micro-diagnostics through constrained zero-shot performance. Meanwhile, even minimal fine-tuning leads to a dramatic performance gain of AgroNVILA on AgMMU, strongly demonstrating its generalization capability empowered by AgroOmni. Full training scripts are publicly available at \url{https://github.com/Jrookie2005/AgroOmni}, and the AgroOmni dataset is released at \url{https://huggingface.co/datasets/AgroOmni/AgroOmni}.
    % Extensive experiments demonstrate that AgroNVILA outperforms state-of-the-art MLLMs, achieving significant improvements(\textbf{+15.03\%}) in multi-altitude agricultural reasoning, %with a new state-of-the-art on the **AgroMind benchmark**, 
    % reflecting its robust capability for holistic agricultural spatial planning.
\end{abstract}

\section{Introduction}
\label{sec:intro}

% 第一段农业大模型很重要，随着MLLM的发展给农业领域带来了非常大的空间之类的，然后也有很多优秀的工作（可以举例列一下，语言类的还有多模态类的）这一段只讲重要意义和发展很快
Agriculture serves as the foundational cornerstone of global food security and economic stability. ~\cite{worldbank_agriculture_food,unicef2024state,alexopoulos2023complementary}
%补充传统农业领域的难点，为什么要用AI
However, agricultural management faces severe bottlenecks due to its inherent complexity. First, effective agronomic decision-making requires a vast knowledge base (ranging from meteorology to plant pathology), yet the limited availability of agricultural experts cannot scale to meet the needs of a massive, globally distributed user base of farmers.~\cite{yang2022transforming} Second, modern farming environments generate rich, highly heterogeneous data across multiple modalities and scales—from Ground level natural images to top-down Unmanned Aerial Vehicle(UAV) and satellite imagery. Human is fundamentally incapable of processing this immense volume of multi-source data efficiently. Consequently, Artificial Intelligence (AI), particularly Large Multimodal Models (MLLMs), has emerged as an inevitable paradigm shift.~\cite{worldbank_agriculture_food,harnessing,zhai2020decision} AI offers the unique capability to democratize expert-level reasoning, fuse complex heterogeneous data, and deliver scalable, precise interventions to millions of users. 

In recent years, the rapid evolution of artificial intelligence has introduced a promising paradigm shift toward autonomous precision farming \cite{harnessing, mllmsurvey, leveraging, agrigpt, agrollm, agrigpt_vl, agrigpt_omni, agrifm,agriworld}. Recent pioneering works have significantly expanded the technological ecosystem. Foundational architectures have rapidly evolved from language-centric agricultural assistants \cite{agrigpt, shizishangpt, agrollm} to specialized multimodal and remote-sensing foundation models \cite{agrigpt_vl, agrigpt_omni, agrifm, cropsts,cao2026crossearthgate}, and even sophisticated agentic execution environments \cite{agriworld,ye2026agrotools}. Concurrently, to evaluate and fuel these advanced reasoning engines, the community has made initial strides in constructing domain-specific training corpora (\textit{e.g.}, AgroInstruct \cite{agrogpt} and AGBASE \cite{agmmu}) and establishing rigorous evaluation benchmarks (\textit{e.g.} AgroMind \cite{li2025agromind}, AgMMU \cite{agmmu}, AgroBench \cite{agrobench}). 

% Together, these synergistic advancements in model architectures, datasets, and evaluation protocols highlight a vibrant and rapidly expanding frontier in AI-driven agriculture.

% 第二段再讲现有的工作有什么缺点，主要聚焦于单个视角，但是农业很多场景是大尺度的考虑，需要uav、satellite这些数据的引入，现有的这些模型没有考虑这些多视角的问题
% 这段不够深入，而且第一句和最后一句实际上表达的意思是一样的；得先解释什么是micro-level和macro-level
Despite this rapid progress, a critical limitation is that existing agricultural MLLMs \cite{agrigpt_vl, agrogpt} rely almost exclusively on close-up, Ground imagery designed for micro-level diagnostics, such as identifying individual leaf diseases or isolated pest features.
However, as mentioned above, modern precision agriculture also operates on a macro scale, requiring multi-altitude observations from Unmanned Aerial Vehicle (UAV) and Satellite to capture plot-level crop dynamics, irrigation patterns, and large-scale field layouts. 
Therefore, bridging the gap between micro-level diagnostics and macro-level spatial analysis through a unified multi-view MLLM is a fundamental requirement for achieving truly holistic agricultural intelligence.

% Current agricultural MLLMs \cite{agrigpt_vl, agrogpt} generally overlook these multi-view requirements, leaving a significant void in holistic agricultural intelligence.

% 第三段讲要做多视角的农业大模型会有什么挑战？首先第一个是数据上的挑战，虽然现在有了评测数据，但是还没有训练数据？第二个是方法上的挑战，做多视角的训练会有什么问题，这个问题最好是能跟你的方法和钧淇的方法可以贴合上的，形成call back
% c) 挑战：数据挑战 + 方法挑战
% 数据集缺乏：收集困难，质量控制等
The development of a multi-view agricultural MLLM is primarily hindered by the absence of multi-view, high-quality VQA datasets. Building a training corpus for a multi-view, cross-scale agricultural VQA dataset presents two fundamental challenges. 
(1) \textbf{Extreme diversity of data sources}, requires the careful aggregation of highly heterogeneous datasets across diverse sources, scenarios, and task categories. It necessitates complex processing and manual screening to align inconsistent formats and annotations into a unified standard. (2) \textbf{Weak semantic relevance of raw agricultural annotations} necessitates intensive instruction engineering to bridge the gap between basic recognition and expert reasoning. Most raw annotations must be reverse-engineered into agricultural VQA pairs that translate low-level visual evidence into agronomic decisions. This process requires manual question design to ensure that conversational intelligence remains strictly consistent with agricultural scene.

% First, from a data perspective, there is an acute scarcity of multi-view instruction-tuning corpora. While pioneering benchmarks like AgroMind \cite{li2025agromind} have established rigorous multi-view evaluation protocols, the community lacks the corresponding training fuel. Existing training datasets are fundamentally focusing on ground level. 

% For instance, as shown in Table~\ref{tab:dataset_comparison}, massive vision-language datasets  (e.g., the 3M+ Agri-3M-VL \cite{agrigpt_vl}) and expert-tuned instruction sets (e.g., the 70K AgroInstruct \cite{agrogpt}, the 50K AGBASE \cite{agmmu}) are predominantly anchored in Ground observations. A unified corpus that explicitly aligns cross-scale observations (Ground, UAV, Satellite) for conversational intelligence remains conspicuously absent. 

% Second, from a methodological perspective, simply feeding multi-view data into standard MLLMs leads to severe architectural bottlenecks. On the perception side, models suffer from {scale and perspective confusion}, erroneously interpreting macro-level field textures as micro-level leaf structures due to their inherent terrestrial bias. 
% On the reasoning side, the extreme task heterogeneity causes {logic drift}, where models rely on statistical shortcuts rather than genuine agronomic reasoning.
\begin{figure}[t]
  \centering
  \includegraphics[width=\linewidth]{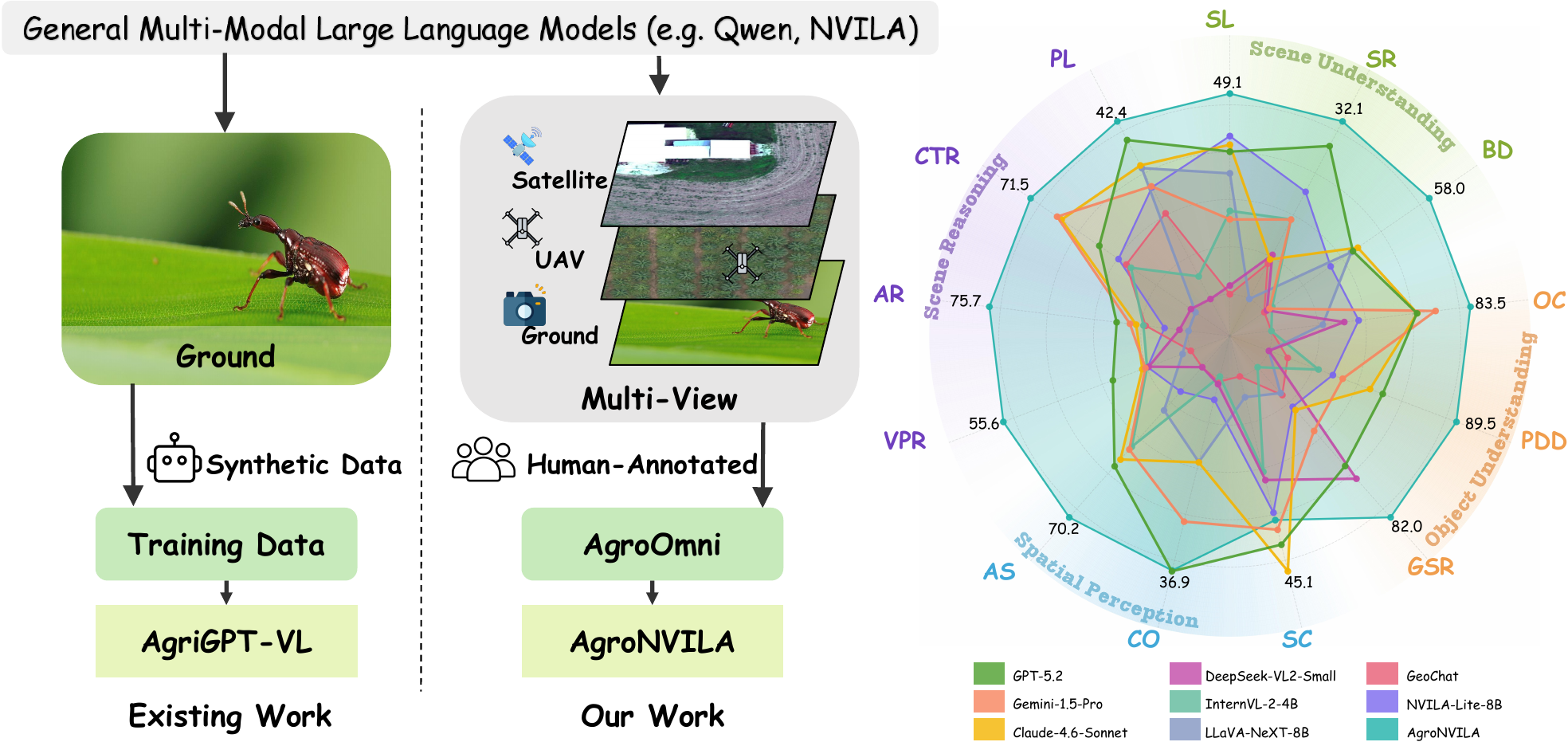}
    \caption{\textbf{Overview} Comparison of our framework with existing paradigms, and its comprehensive performance radar on the AgroMind benchmark (numbers give the best score per task among the models shown)}
  \label{fig:Teaser}
  \vspace{-1em}
\end{figure}
% Our Solution: 
% 因此，我们首先提出了一个数据集,我们制作了训练集AgroOmni
% 针对上面提到的困难，我们怎么怎么样做了这个数据集
To overcome the data bottleneck, we introduce \textbf{AgroOmni}, a pioneering, large-scale multi-modal instruction-tuning dataset. AgroOmni seamlessly integrates Ground, UAV, and Satellite perspectives, providing a comprehensive ``multi-altitude'' context. It comprises 268,119 professionally curated QA pairs, bridging the semantic gap by transforming raw labels into semantically rich, expert-level agronomic reasoning chains spanning 14 specialized task types. By expanding beyond simple identification to encompass complex top-down spatial reasoning and planning, AgroOmni serves as the essential domain-specific fuel for multi-view agricultural intelligence.

To validate the effectiveness of the AgroOmni dataset, we establish a robust baseline, \textbf{AgroNVILA}, and explore targeted methods designed for complex multi-view agricultural scenarios. By training on AgroOmni, our model achieves large and view-asymmetric gains on the UAV and Satellite views. This localized leap in accuracy on UAV and Satellite perspectives effectively proves that high-quality, cross-scale data is the fundamental prerequisite for rectifying the \textit{Ground-level bias} and achieving comprehensive agricultural reasoning. We make all training scripts and the AgroOmni dataset publicly available at \url{https://github.com/Jrookie2005/AgroOmni} and \url{https://huggingface.co/datasets/AgroOmni/AgroOmni}
In summary, our key contributions are as follows:

\begin{itemize}
    \item We formally identify the Ground view bias and the resulting scale confusion in current MLLMs. We define multi-view spatial reasoning, bridging Ground, UAV, and Satellite perspectives as a crucial new challenge for achieving comprehensive agricultural intelligence.
    \item We curate \textbf{AgroOmni}, a large-scale (268,119), multi-view agricultural instruction-tuning dataset. Spanning 56 expert-level task categories, it serves as the essential domain-specific fuel to correct the spatial cognitive blind spots of existing models.
    \item We establish a robust baseline, \textbf{AgroNVILA}, tailored for multi-view agricultural scenarios. It achieves a new state-of-the-art (59.86\%) on the AgroMind benchmark, significantly outperforming GPT-5.2 (47.29\%) and shows great adaption on AgMMU.
\end{itemize}

\section{Related Work}
\label{sec:related}
% Multi-modal Large Language Models: 简述 LLaVA, NVILA 等基座模型。 
\textbf{General Multi-modal Large Language Models.}
Pioneering Multi-modal Large Language Models (MLLMs), such as LLaVA \cite{llava1.5} and BLIP-2 \cite{blip2}, established the standard modular architecture by aligning a pre-trained vision encoder with an LLM. Recent advancements have significantly scaled this paradigm. Open-weights models (e.g., LLaVA-NeXT \cite{llavanext}, InternVL \cite{internvl}, NVILA \cite{nvila}) utilize dynamic resolution strategies (e.g., AnyRes) to handle high-resolution inputs, while proprietary models (e.g., GPT-4o \cite{gpt4o}, Gemini-1.5-Pro \cite{gemini1.5}) exhibit remarkable zero-shot reasoning across general visual question answering and instruction following domains~\cite{zheng2026weathersyn,cao2026taskadaptive}. Despite these achievements, current state-of-the-art MLLMs suffer from a critical \textit{Ground-level bias}. Because their foundational vision encoders and alignment corpora are overwhelmingly dominated by internet-crawled, human-perspective images, their internal spatial representations are inherently skewed.

% AI in Agriculture: 简述现有农业多模态模型发展情况。
\textbf{Multi-modal Large Language Models in Agriculture.}
MLLMs signify a paradigm shift in agricultural AI, transitioning from isolated visual recognition to complex agronomic reasoning \cite{harnessing, agrigpt_vl, agrillava, agrogpt}. Early foundational datasets primarily targeted narrow perception tasks, such as pest and disease identification \cite{plantvillage,ip102}. Recently, domain-specific training corpora Table~\ref{tab:dataset_comparison} (e.g., AgroInstruct \cite{agrogpt}, AGBASE \cite{agmmu}, Agri-3M-VL \cite{agrigpt_vl}) and rigorous benchmarks (e.g., AgMMU \cite{agmmu}, AgroBench \cite{agrobench}, AgroMind \cite{li2025agromind}, AgroVG \cite{li2026agrovg}) have emerged to cultivate broader vision-language capabilities. Consequently, generic architectures (e.g., LLaVA-OneVision \cite{llava_onevision}, Qwen-VL \cite{qwen3vl}, NVILA \cite{nvila}) have been adapted into specialized MLLMs, such as AgriGPT-VL \cite{agrigpt_vl} and Agri-LLaVA \cite{agrillava}, excelling in fine-grained diagnosis and visual question answering. However, despite this expansion, existing resources remain predominantly restricted to ground-level perception. They lack the large-scale multi-view coverage (e.g., UAV and Satellite view) required for 
%holistic 
top-down planning, resulting in fragmented, Ground-level efforts rather than a cohesive, cross-scale spatial foundation.
\begin{table}[t]
\centering
\small
\caption{\textbf{Comparison of existing agricultural vision-language instruction datasets.} \textit{SFT QA} refers to the number of complex instruction-tuning samples used for the fine-tuning stage. \textit{Grounding} strictly indicates support for region-level geometric localization (e.g., bounding boxes). \textit{Multi-View} denotes the inclusion of macroscopic perspectives (UAV/Satellite) alongside standard Ground imagery.}
\label{tab:dataset_comparison}
\begin{tabular}{@{}l c c c c c@{}}
\toprule
\textbf{Dataset} & Images & \textbf{SFT QA} & \textbf{Complex Reasoning} & \textbf{Spatial Grounding} & \textbf{Multi-View} \\
% \multirow{2}{*}{\textbf{Dataset}} & \multirow{2}{*}{\textbf{Images}} & \textbf{SFT} & \textbf{Complex} & \textbf{Spatial} & \textbf{Multi-} \\
 % & & \textbf{QA} & \textbf{Reasoning} & \textbf{Grounding} & \textbf{View} \\
\midrule
AgroInstruct \cite{agrogpt} & 108K & 70K & $\times$ & $\times$ & $\times$ \\
Agri-LLaVA \cite{agrillava} & 397K & 6K* & $\times$ & $\times$ & $\times$ \\
Agri-3M-VL \cite{agrigpt_vl} & 1M & 3M & \checkmark & $\times$ & $\times$ \\
AGBASE \cite{agmmu} & 57K & 57K & \checkmark & $\times$ & $\times$ \\
\textbf{AgroOmni (Ours)} & \textbf{98K} & \textbf{268K} & \textbf{\checkmark} & \textbf{\checkmark} & \textbf{\checkmark} \\
\bottomrule
\multicolumn{6}{l}{\small *Agri-LLaVA claims 400K data, but only 6K high-quality dialogues are used for SFT.}
\end{tabular}
\vspace{-1.5em}
\end{table}

\section{AgroOmni}
\label{sec:agroomni}

%To dismantle the Ground-level bias inherent in existing agricultural models, we introduce \textbf{AgroOmni}, a comprehensive, multi-view instruction-tuning corpus.

\begin{figure}[tb]
  \centering
  \includegraphics[width=\textwidth]{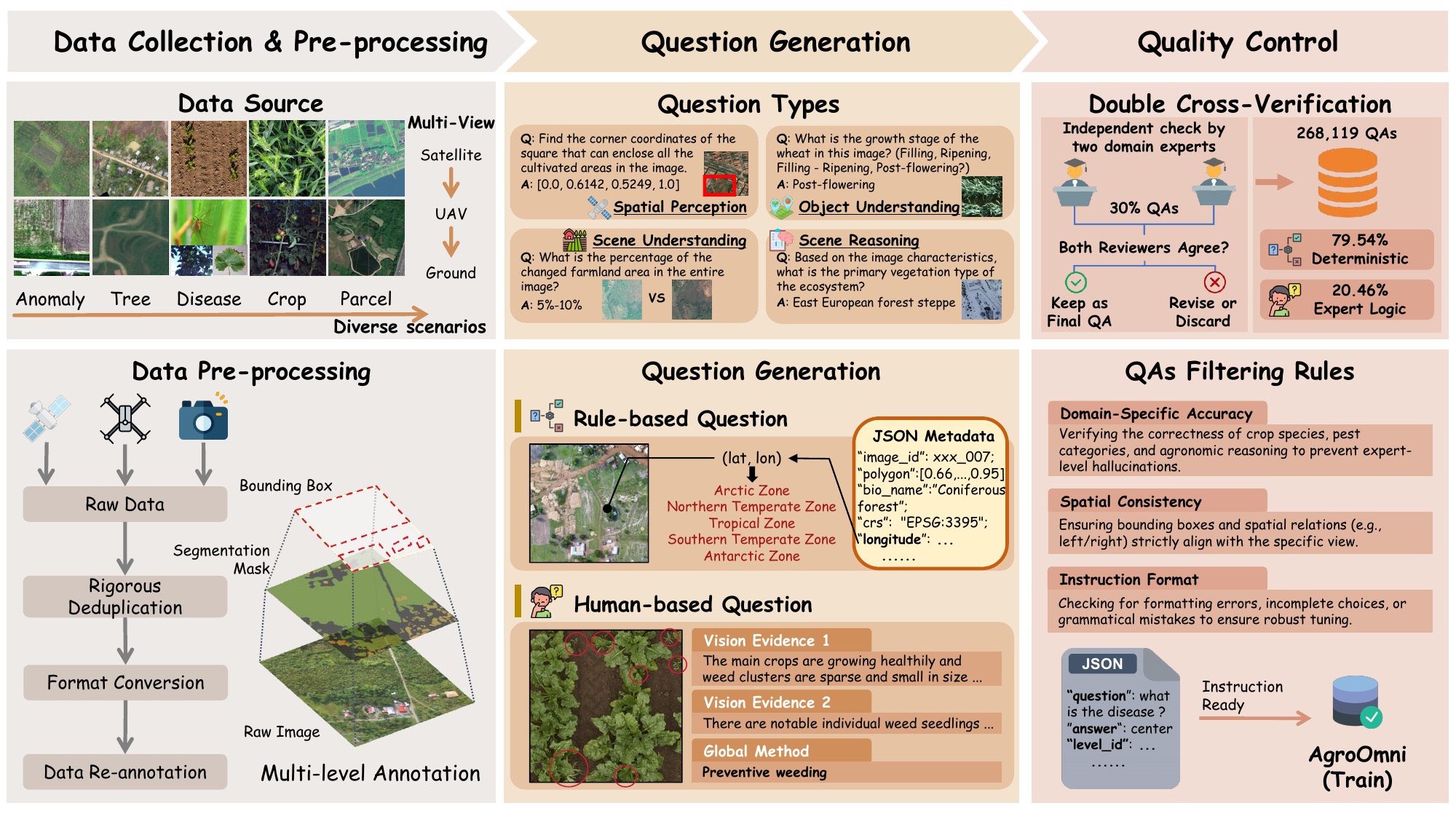}
  \caption{The \textbf{curation pipeline of AgroOmni} training set contains three stages, i.e., data collection and pre-processing, question generation, and quality control.}
  \label{fig:data_pipeline}
  \vspace{-2em}
\end{figure}

%\subsection{Data Construction}
% This section details our pipeline from data collection to question generation.

\textbf{Data Collection.}
As shown in Fig.~\ref{fig:data_pipeline}, we aggregate data from three heterogeneous sensor modalities (Ground, UAV, and Satellite) to encompass the full spatial scale of agricultural production. Our \textbf{AgroOmni} consolidates \textbf{97,926} images from a proprietary remote sensing parcel collection and 12 public datasets, including CLCD~\cite{liu2022CLCD}, EarthVQA~\cite{wang2024earthvqa}, GWHD~\cite{David2021Wheat}, OAM-TCD~\cite{veitchmichaelis2024oamtcdgloballydiversedataset}, OilPalmUAV~\cite{zheng2021growing}, AVC~\cite{Chiu_2020_CVPR}, PhenoBench~\cite{weyler2024pami}, IP102~\cite{ip102}, ACFR~\cite{bargoti2016deep}, Tomato Dataset~\cite{zhang2025Tomato}, 2018 AI Challenge Dataset, and CropHarvest~\cite{tseng2021cropharvest} (see details in Appendix~\ref{suppl:dataset}). In Table~\ref{tab:dataset_comparison}, AgroOmni advances the state-of-the-art by uniquely integrating multi-view perspectives and spatial grounding into instruction-tuning. 
% Notably, the integration of multi-temporal sources (e.g., GWHD~\cite{David2021Wheat}, CLCD~\cite{liu2022CLCD}) effectively addresses the \textit{Temporal Variation} task gap. 
% To prevent data leakage, we enforce strict physical separation from AgroMind~\cite{li2025agromind} with zero image-level overlap. 
Furthermore, our proprietary parcel collection and multi-temporal sources (\textit{e.g.}, the GWHD~\cite{David2021Wheat}) significantly broaden the scope toward {Temporal Variation} (TV) tasks beyond existing scenarios of Agromind \cite{li2025agromind}.

\textbf{Data Pre-processing.}
We adopt and extend the rigorous data processing pipeline established in AgroMind~\cite{li2025agromind}, implementing customized protocol to handle the heterogeneity of our data sources. This protocol involves multi-stage cleaning, including manual screening to remove defective samples and content-aware format conversion. We further apply randomized parcel-boundary cropping~\cite{zheng2026comprehensive} to generate multi-scale samples, simulating diverse spatial query variations. All original image resolutions are preserved to maintain data fidelity. Detailed collection procedures and standardized preprocessing protocols are documented in Appendix~\ref{suppl::preprocessing} and~\ref{suppl::detail data}.

\textbf{Question Generation.}
As depicted in Fig.~\ref{fig:data_pipeline}, we utilize a Dual-Track generation method to transform hierarchical annotations into instruction-tuning data.
\textbf{Rule-based Generation} focuses on tasks demanding rigorous numerical and positional accuracy by employing a deterministic template filling strategy. We utilize extracted JSON metadata and segmentation masks to directly map precise polygon coordinates or geographical attributes into parameterized queries. This strict reliance on absolute ground truth effectively eliminates the numerical and spatial hallucinations frequently exhibited by current vision-language models.
\textbf{Evidence-based Logic Synthesis} for complex reasoning tasks adopts a ``reverse-engineering'' approach, translating raw labels into dynamic reasoning chains. By requiring the model to explicitly identify visual evidence before deriving a final conclusion, this strategy successfully balances semantic richness with factual accuracy while mitigating risks associated with open-ended generation. Further details are provided in Appendix~\ref{suppl::Question Generation}.
% The foundation of robust spatial reasoning relies on the heterogeneity of visual inputs. We systematically constructed the raw image reservoir across three distinct spatial scales: Ground-view (micro-level crop phenotypes), UAV-view (intermediate-scale plot density), and Satellite-view (macro-scale regional planning). 

\textbf{Quality Control.}
To guarantee the integrity of the instruction-tuning corpus, we employ a straightforward quality screening process (as illustrated in Fig.~\ref{fig:data_pipeline}). During the generation phase, we iteratively filter and regenerate QA pairs that exhibit broken formats, unparsable syntax, or incomplete choices. We also conduct basic spatial and semantic checks, discarding samples where bounding boxes clearly mismatch the visual content or agronomic terms deviate from the source metadata. For benchmark decontamination, we remove duplicates with byte-exact hashing, footprint overlap, and perceptual matching, and we group samples by parcel, sequence, and geolocation. These steps enforce a strict zero-overlap boundary with the AgroMind evaluation set (Appendix~\ref{suppl::preprocessing}).

\begin{figure}[tbb]
  \centering
  \includegraphics[width=1.02\linewidth, keepaspectratio=true]{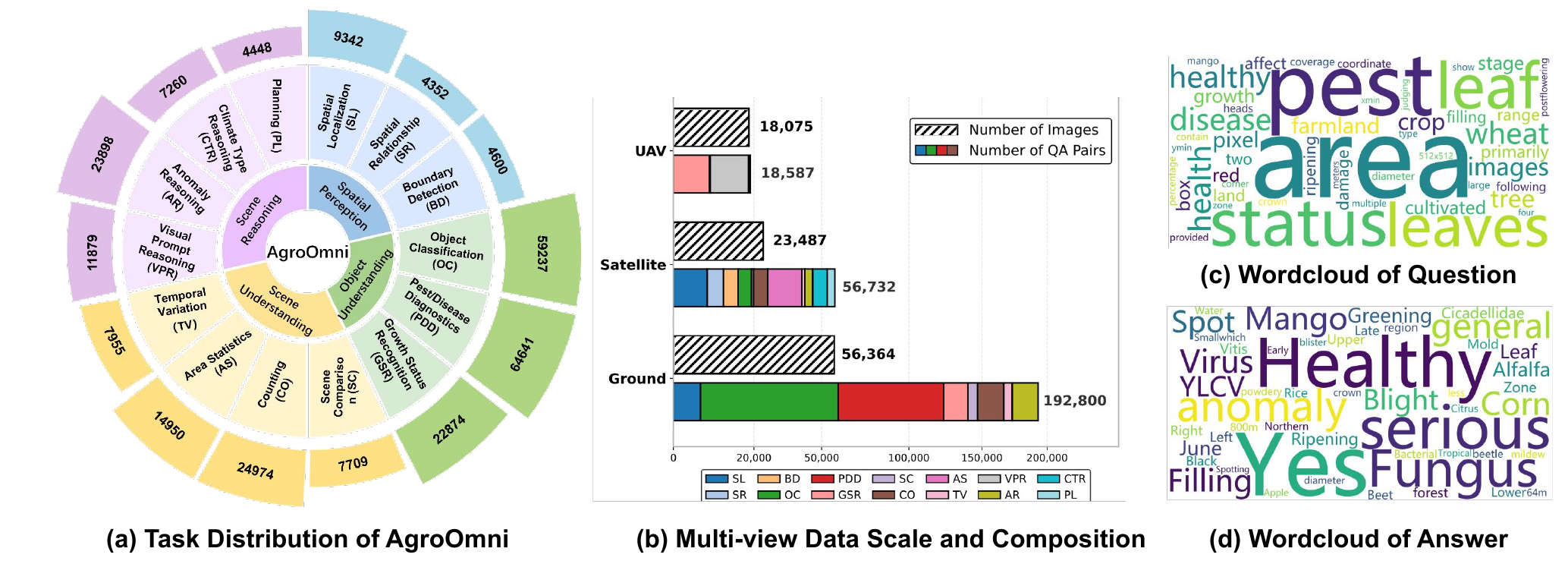}
  \caption{\textbf{Comprehensive statistics of the AgroOmni.} (a) Classification of 14 fine-grained agricultural tasks across four cognitive dimensions (b) Multi-view data scale and task distribution across UAV, Satellite, and Ground perspectives  (c-d) Wordclouds for QA pairs, highlighting the high density of domain-specific agronomic terminology}
  \label{fig:data_statistics}
  \vspace{-1.5em}
\end{figure}
 
\section{Dataset Statistics and Analysis}

AgroOmni is structured with a balance between expansive coverage and cognitive depth. As visualized in Fig.~\ref{fig:data_statistics}, the dataset's characteristics are three-fold:

\textbf{Heterogeneous Distribution.} 
As shown in Fig.~\ref{fig:data_statistics}(b), AgroOmni aggregates \textbf{268,119} QA pairs across three distinct views. Ground-level imagery (71.9\%) constitutes the majority, capturing micro-scale crop phenotypes and fine-grained pathological features essential for disease and pest diagnosis. Satellite imagery (21.2\%) provides macroscopic contexts required for regional land coverage analysis, boundary delineation, and climate reasoning. Bridging these two extremes, the UAV perspective (6.9\%) captures meso-scale spatial topologies necessary for plot-level plant density estimation and weed-crop interaction analysis. 

Notably, the dataset exhibits a native extremes-dominant distribution. This skew is an objective reflection of real-world agricultural data acquisition, where foundational perception tasks naturally outnumber complex planning scenarios. We intentionally preserve this authentic heterogeneity rather than artificially discarding useful samples. Training on such an unbalanced corpus serves as a rigorous stress test, challenging models to internalize cross-scale agronomic logic rather than relying on statistical shortcuts. This inherent distribution challenge underscores a critical research frontier: the need for advanced spatial anchoring mechanisms and adaptive learning strategies that can maintain logical consistency across highly skewed agricultural task domains.

\textbf{Hierarchical Task Taxonomy.} 
Fig.~\ref{fig:data_statistics}(a) delineates 14 task types, mapping a cognitive progression from fundamental Spatial Perception to sophisticated Scene Reasoning. The foundation of the task taxonomy is \textbf{Spatial Perception}, consisting of Spatial Localization, Spatial Relationship, and Boundary Detection. This tier enforces fundamental geometric grounding, ensuring that models accurately anchor coordinates and topological layouts before proceeding to semantic tasks. Building upon this geometric baseline, \textbf{Object Understanding} targets fine-grained semantic recognition through Object Classification, Pest/Disease Diagnosis, and Growth Status Recognition. Serving as the most data-intensive tier, it evaluates the model's capacity to extract individual morphology and pathological features, mirroring the predominant demand for daily field diagnostics. Furthermore, the paradigm elevates from isolated entities to holistic contexts in \textbf{Scene Understanding}. This tier encompasses Counting, Area Statistics, Scene Comparison, and Temporal Variation. By integrating spatial quantification with Temporal Variation, this level mandates the model to move beyond static observation to interpret quantitative distributions and track longitudinal land-use dynamics. Finally, the hierarchy culminates in \textbf{Scene Reasoning}, which synthesizes all preceding perceptual inputs for expert-level decision-making. Through Visual Prompt Reasoning, Anomaly Reasoning, Climate Type Reasoning, and Planning, this apex tier compels the model to execute complex multi-step deductions and formulate agronomic interventions based on diverse multi-scale observations.

\textbf{Lexical Analysis.} 
Lexical analysis in Fig.~\ref{fig:data_statistics}(c, d) demonstrates high domain specificity. The question distribution (Fig.~\ref{fig:data_statistics}(c)) reveals a strong emphasis on targeted, analytical inquiries. Dominant tokens such as \textit{area}, \textit{box}, and \textit{pixel} underscore the dataset's rigorous demand for geometric grounding and spatial quantification. Concurrently, condition-oriented tokens like \textit{status}, \textit{disease}, and \textit{leaf} explicitly direct the model's attention toward agronomic monitoring rather than generic visual description. Correspondingly, the answer distribution (Fig.~\ref{fig:data_statistics}(d)) transitions completely from conversational vocabulary to strict professional taxonomy. The outputs are densely populated with precise biological classifications (\textit{Fungus}, \textit{Blight}, \textit{Virus}, \textit{YLCV}), definitive diagnostic judgments (\textit{Healthy}, \textit{serious}, \textit{anomaly}), and exact phenological or environmental labels (\textit{Filling}, \textit{Tropical}, \textit{Zone}). By confining the output space to rigorous agricultural terminology, AgroOmni forces models to directly map complex visual evidence to concise, expert-aligned reasoning.

\section{Experiment}
\label{sec:experiment}
\subsection{Experimental Setup} 
% 各项农业 Benchmarks、评价指标。

% \textbf{Baseline Models.} 

\textbf{AgroNVILA} To validate the effectiveness and practicality of the AgroOmni dataset, we adopt NVILA-Lite-8B with a frozen SigLIP vision encoder as our base architecture. We first perform SFT on the 268,119 QA pairs of AgroOmni to establish a robust baseline. 
% The SFT stage is optimized using a cosine decay learning rate peaking at $2 \times 10^{-5}$ and a global batch size of 32 for one epoch. Further experimental and reproducibility details are deferred to Appendix~\ref{suppl:reproducibility}.
To further demonstrate the dataset's capacity to support diverse alignment paradigms, We subsequently apply three standard Reinforcement Learning (RL) algorithms on the SFT baseline: Proximal Policy Optimization (\textbf{PPO}) \cite{PPO}, Group Relative Policy Optimization (\textbf{GRPO}) \cite{GRPO}, and Domain-aware Relative Policy Optimization (\textbf{DRPO}) \cite{DRPO}. Comprehensive details regarding the RL preliminaries, task-specific reward formulations, and KL-aware regularization strategies are provided in Appendix~\ref{suppl:method}. 
% During the RL stage, the models are trained for one epoch with a learning rate of $1 \times 10^{-6}$ for the LLM and $1 \times 10^{-7}$ for the projector. For group-based methods, we set the group size to 8 for advantage computation. All experiments employ a gradient accumulation of 4 steps and are conducted on 4$\times$A100 GPUs. 
Overall, AgroNVILA is ultimately constructed via SFT initialization followed by GRPO alignment. Further experimental details are deferred to Appendix~\ref{suppl:reproducibility}.
% Further experimental details are deferred to Appendix~\ref{suppl:reproducibility}.

\textbf{Evaluation.}
We evaluate our model on the AgroMind benchmark \cite{li2025agromind}, an agricultural VQA benchmark covering Ground, UAV, and Satellite imagery with 13 tasks across four dimensions: Spatial Perception, Object Understanding, Scene Understanding, and Scene Reasoning. The MLLMs evaluated on AgroMind are grouped into three categories: closed-source models, open-source MLLMs, and our variants. Following the official protocol, we report per-task accuracy (\%), and we take the overall score as the micro-average over all evaluation questions. 

\textbf{Models.} 
To establish a rigorous and comprehensive evaluation baseline, we curate a diverse suite of MLLMs representing distinct paradigms. First, we sample representative open-weight models from the original AgroMind~\cite{li2025agromind} benchmark, encompassing both general-purpose architectures (e.g., LLaVA-NeXT~\cite{llavanext}, InternVL-2~\cite{internvl2}) and spatially-adapted models (e.g., GeoChat~\cite{geochat}). Second, to probe the absolute zero-shot upper bound of generic world knowledge, we augment the evaluation with the latest state-of-the-art proprietary models, specifically GPT-5.2 and Claude-4.6-Sonnet. Third, we include AgriGPT-VL~\cite{agrigpt_vl} as a domain-specific reference to contextualize prior efforts in agricultural adaptation. Finally, we evaluate our proposed AgroNVILA, which is derived by applying our complete alignment pipeline (SFT and GRPO) to the base NVILA-Lite-8B~\cite{nvila} architecture. This structured taxonomy systematically isolates the performance delta between generic pre-training, existing domain patches, and our high-density alignment strategy. More details are in Appendix~\ref{suppl:agromind_eval_details}.

\subsection{Main Results}
\label{sec:main_results}

\textbf{Comparison with State-of-the-Art MLLMs on AgroMind.}
Table~\ref{tab:overall_performance} presents a comprehensive performance comparison of 17 leading MLLMs against our foundational baseline trained on the AgroOmni corpus. Strikingly, by merely performing supervised fine-tuning on our dataset, AgroNVILA achieves a state-of-the-art average accuracy of 59.52\% on the AgroMind benchmark (59.86\% after RL alignment), directly validating the quality and necessity of this multi-view corpus.
Despite their immense parameter scale, leading proprietary models exhibit severe performance degradation on agricultural tasks. For instance, GPT-5.2 and Gemini-1.5-Pro only achieve overall accuracies of 47.29\% and 41.06\%, respectively. In rigorous geometric tasks that require multi-altitude spatial grounding, our AgroNVILA significantly outperforms GPT-5.2 by absolute margins of +15.93\% in Boundary Detection (BD) and +16.91\% in Area Statistics (AS). 
Similar performance gaps are observed within the open-source and specialized ecosystems. Standard foundational models like LLaVA-NeXT-8B (31.53\%) and remote sensing experts like GeoChat (25.93\%) struggle significantly across the benchmark. Specifically, in complex Scene Reasoning tasks such as Anomaly Reasoning (AR), AgroNVILA secures a 75.74\% accuracy, decisively outperforming both LLaVA-NeXT-8B (17.22\%) and GeoChat (29.80\%). These quantitative results conclusively demonstrate that training on AgroOmni effectively bridges the performance gap across heterogeneous, cross-scale agricultural scenarios.

% 与 SOTA 模型（LLaVA-1.5, NVILA, Qwen-VL 等）的横向对比大表。on AgroMind
\begin{table}[t]
\centering
\setlength{\tabcolsep}{2.5pt} % 调整列间距
\renewcommand{\arraystretch}{1.0} % 调整行高
\scriptsize
% \caption{\textbf{Performance comparison.} The overall score represents the average accuracy across all tasks. Task names are abbreviated for brevity. \textbf{Bold} and \underline{underlined} values denote the best and second best results within each column respectively.}
\caption{\textbf{Performance comparison on AgroMind benchmark.} The overall score is the micro-average over all evaluation questions. \textbf{Bold} and \underline{underlined} values denote the best and second-best results respectively. Task abbreviations: AR (Anomaly Reasoning), AS (Area Statistics), BD (Boundary Detection), CO (Counting), CTR (Climate Type Reasoning), GSR (Growth Stage Recognition), OC (Object Classification), PDD (Pest/Disease Diagnosis), PL (Planning), SC (Scene Comparison), SL (Spatial Localization), SR (Spatial Relationship), and VPR (Visual Prompt Reasoning).}
\label{tab:overall_performance}
\resizebox{1.0\linewidth}{!}{
\begin{tabular}{@{}c ccc ccc ccc cccc c@{}}
\toprule
\multirow{3}{*}{\textbf{Model}} & \multicolumn{3}{c}{\textbf{\begin{tabular}[c]{@{}c@{}}Spatial\\ Perception\end{tabular}}} & \multicolumn{3}{c}{\textbf{\begin{tabular}[c]{@{}c@{}}Object\\ Understanding\end{tabular}}} & \multicolumn{3}{c}{\textbf{\begin{tabular}[c]{@{}c@{}}Scene\\ Understanding\end{tabular}}} & \multicolumn{4}{c}{\textbf{\begin{tabular}[c]{@{}c@{}}Scene\\ Reasoning\end{tabular}}} & \multirow{3}{*}{\textbf{Overall}} \\
\cmidrule(lr){2-14}
 & \textbf{SL} & \textbf{SR} & \textbf{BD} & \textbf{OC} & \textbf{PDD} & \textbf{GSR} & \textbf{SC} & \textbf{CO} & \textbf{AS} & \textbf{VPR} & \textbf{AR} & \textbf{CTR} & \textbf{PL}  & \\
\midrule
Human & 19.15 & 15.00 & 40.83 & 57.60 & 46.17 & 47.22 & 35.49 & 28.88 & 39.46 & 12.33 & 25.62 & 34.07 & 13.50 & 33.15 \\
Random & 18.03 & 17.97 & 16.48 & 35.65 & 25.65 & 10.00 & 24.39 & 4.99 & 21.20 & 9.19 & 19.87 & 21.70 & 26.96 & 19.24 \\
\midrule
GPT-4o~\cite{gpt4o} & 41.38 & \textbf{35.16} & 33.52 & \underline{78.55} & \underline{66.09} & 55.83 & 43.90 & 23.75 & 46.20 & 16.10 & 27.15 & \underline{69.81} & 30.90 & 43.14 \\
GPT-5.2 & 39.19 & 29.94 & 42.09 & 74.14 & 65.61 & 60.50 & 41.22 & \textbf{36.85} & \underline{53.27} & \underline{28.09} & 38.33 & 57.41 & \underline{38.83} & \underline{47.29} \\
Gemini-1.5-Pro~\cite{gemini1.5} & 27.59 & 23.44 & 24.73 & 77.39 & 52.61 & 45.83 & 39.02 & 32.30 & 47.78 & 19.92 & 34.44 & 66.04 & 30.04 & 41.06 \\
Claude-3.5-Sonnet & 24.14 & 17.19 & 40.11 & 57.97 & 50.00 & 57.50 & \underline{44.51} & 24.47 & 35.44 & 20.34 & 33.11 & 50.94 & 36.05 & 37.11 \\
Claude-4.6-Sonnet & 40.37 & 19.91 & \underline{43.13} & 74.20 & 61.48 & 37.08 & \textbf{45.14} & 26.87 & 51.07 & 20.69 & 32.61 & 65.14 & 34.00 & 43.15 \\
\midrule
DeepSeek-VL2-small~\cite{deepseekvl} & 16.26 & 20.31 & 24.18 & 61.45 & 28.70 & \underline{65.83} & 31.71 & 19.71 & 20.57 & 19.07 & 19.87 & 38.68 & 8.58 & 28.08 \\
TinyLLaVA~\cite{tinyllava} & 29.56 & 17.97 & 24.73 & 68.41 & 23.48 & 30.00 & 24.39 & 23.99 & 31.65 & 3.39 & 13.91 & 45.28 & 9.44 & 28.01 \\
InternVL-2-4B~\cite{internvl2} & 29.06 & 23.44 & 25.27 & 48.70 & 44.78 & 19.17 & 30.49 & 19.00 & 46.84 & 19.49 & 30.46 & 50.94 & 12.88 & 31.15 \\
XComposer2-4K-HD~\cite{xcomposer} & 29.06 & 27.34 & 32.97 & 51.88 & 40.00 & 44.17 & 34.15 & 22.33 & 27.53 & 11.86 & 37.75 & 43.40 & 38.63 & 33.02 \\
InstructBLIP-Vicuna-7B~\cite{instructblip} & 9.85 & 16.41 & 20.88 & 33.33 & 21.74 & 10.83 & 19.51 & 10.93 & 21.52 & 4.66 & 20.53 & 20.75 & 11.16 & 17.39 \\
Mantis-Idefics2~\cite{mantis} & 16.75 & 20.31 & 31.32 & 63.77 & 44.35 & 37.50 & 22.56 & 24.23 & 20.89 & 8.47 & 13.91 & 56.60 & 30.90 & 30.41 \\
LLaVA-NeXT-7B-Mistral & 33.50 & 22.66 & 40.11 & 44.64 & 40.00 & 36.67 & 21.95 & 24.70 & 36.71 & 8.05 & 15.23 & 39.62 & 11.59 & 29.17 \\
LLaVA-NeXT-8B~\cite{llavanext} & 35.47 & 16.41 & 41.76 & 57.68 & 28.70 & 30.00 & 19.51 & 26.84 & 34.81 & 10.59 & 17.22 & 37.74 & 33.48 & 31.53 \\
GeoChat~\cite{geochat} & 14.78 & 19.53 & 23.63 & 48.70 & 34.78 & 30.83 & 16.46 & 19.48 & 20.57 & 8.47 & 29.80 & 51.89 & 24.89 & 25.93 \\
GeoLLaVA-8K~\cite{geollava8k} & 9.85 & 25.78 & 17.58 & 34.78 & 15.22 & 7.50 & 17.68 & 4.75 & 16.14 & 7.63 & 14.57 & 34.91 & 8.58 & 15.73 \\
AgriGPT-VL~\cite{agrigpt_vl} & 27.41 & 24.73 & 27.53 & 71.18 & 44.61 & 46.08 & 35.43 & 20.66 & 32.49 & 23.94 & \underline{40.63} & 62.35 & 32.12 & 36.97 \\
NVILA-Lite-8B~\cite{nvila} & \underline{41.85} & 25.89 & 37.47 & 63.96 & 49.43 & 35.75 & 36.51 & 21.13 & 28.62 & 19.33 & 24.26 & 53.49 & 29.82 & 35.68 \\
\midrule
AgroNVILA & \textbf{49.12} & \underline{32.12} & \textbf{58.02} & \textbf{83.49} & \textbf{89.52} & \textbf{82.00} & 37.60 & \underline{36.78} & \textbf{70.18} & \textbf{55.63} & \textbf{75.74} & \textbf{71.48} & \textbf{42.38} & \textbf{59.86} \\
\bottomrule
\end{tabular}
}
\vspace{-1em}
\end{table}

\begin{table*}[t]
\centering
\small
\caption{\textbf{Effect of standard RL strategies on the AgroMind benchmark.} We report accuracy (\%) for each task and the overall micro-average over all questions. AgroNVILA (SFT) serves as the starting point for alignment. \textbf{Bold} and \underline{underlined} values denote the best and second-best results within each column respectively.}
\label{tab:rl_ablation}
\resizebox{\textwidth}{!}{%
\begin{tabular}{@{\extracolsep{\fill}}l ccc ccc ccc cccc c@{}}
\toprule
\multirow{2}{*}{\textbf{Method}} & \multicolumn{3}{c}{Spatial Perception} & \multicolumn{3}{c}{Object Understanding} & \multicolumn{3}{c}{Scene Understanding} & \multicolumn{4}{c}{Scene Reasoning} & \multirow{2}{*}{\textbf{Overall}} \\
\cmidrule(lr){2-4} \cmidrule(lr){5-7} \cmidrule(lr){8-10} \cmidrule(lr){11-14}
& \textbf{SL} & \textbf{SR} & \textbf{BD} & \textbf{OC} & \textbf{PDD} & \textbf{GSR} & \textbf{SC} & \textbf{CO} & \textbf{AS} & \textbf{VPR} & \textbf{AR} & \textbf{CTR} & \textbf{PL} & \\
\midrule
AgroNVILA (SFT)   & 49.12 & 31.65 & 55.49 & 83.46 & \textbf{90.00} & \underline{82.25} & \textbf{39.35} & 36.09 & \textbf{70.40} & 52.92 & 74.62 & 70.55 & \textbf{43.53} & 59.52 \\
+PPO \cite{PPO}   & \underline{49.17} & \textbf{32.50} & \underline{58.13} & \textbf{83.72} & 89.65 & \textbf{82.33} & 36.69 & 36.30 & 69.93 & \underline{55.33} & \underline{75.81} & 70.92 & 41.99 & 59.71 \\
+GRPO \cite{GRPO} & 49.12 & 32.12 & 58.02 & 83.49 & 89.52 & 82.00 & \underline{37.60} & \textbf{36.78} & \underline{70.18} & \textbf{55.63} & 75.74 & \textbf{71.48} & \underline{42.38} & \textbf{59.86} \\
+DRPO \cite{DRPO} & \textbf{49.51} & \underline{32.19} & \textbf{58.46} & \underline{83.55} & \underline{89.74} & 81.83 & 37.42 & \underline{36.47} & 70.06 & 54.86 & \textbf{75.87} & \underline{71.30} & 41.99 & \underline{59.77} \\
\bottomrule
\end{tabular}
}%
\end{table*}

\textbf{Impact of Reinforcement Learning Alignment.}
Table~\ref{tab:rl_ablation} demonstrates its compatibility with advanced Reinforcement Learning (RL) alignment paradigms. While standard supervised fine-tuning already establishes a robust foundation (59.52\%), applying policy optimization directly on our dataset consistently elevates overall performance. Specifically, utilizing Group Relative Policy Optimization (GRPO) achieves the highest overall accuracy of 59.86\%, while Domain-aware Relative Policy Optimization (DRPO) closely follows at 59.77\%. The improvements are particularly pronounced in cognitive-heavy dimensions, such as Visual Prompt Reasoning (VPR) and Boundary Detection (BD), where GRPO improves the baseline by +2.71\% and +2.53\%, respectively. These results indicate that the dense agronomic logic and multi-view geometric annotations within AgroOmni provide usable reward signals, although the per-task gains are not uniform. Consequently, the dataset can serve as a practical testbed for future exploration of multimodal alignment in specialized domains.

\textbf{Performance On AgMMU.}
\label{sec:agmmu_performance}
To comprehensively assess the domain generalization capabilities of our AgroNVILA trained on AgroOmni, we evaluate it on AgMMU~\cite{agmmu}, with all visual data solely distilled from mobile phone images of real farmer online inquiries, across five specialized categories: Disease, Insect/Pest, Species, Management, and Symptom.
To ensure a fair comparative analysis, we adopt the official Multiple-Choice Question (MCQ) accuracy metric and strictly retain the evaluation results officially reported by AgMMU~\cite{agmmu}, benchmarking our model against two distinct tiers of state-of-the-art MLLMs: massive proprietary models and leading open-source models operating at a strictly comparable 7B--8B parameter scale. We also supplement this official benchmark suite by evaluating our base architecture, NVILA-Lite-8B~\cite{nvila}.
As detailed in Table~\ref{tab:agmmu_mcq_detailed}, AgroNVILA fine-tuned on merely 30\% of AGBASE achieves an average accuracy of 78.3\%, outperforming all open-source 8B models and approaching the proprietary systems. Beyond the overall scores, the zero-shot baseline (68.2\%) underscores the extreme visual heterogeneity between Ground-level agricultural features and macro-scale UAV/Satellite views. Conversely, exposing the model to just 10\% of AGBASE triggers a massive gain of +8.3\%, demonstrating that AgroNVILA adapts swiftly to a new agricultural corpus. 

\begin{table*}[t]
\centering
\caption{Detailed performance comparison on the AgMMU~\cite{agmmu} benchmark (MCQs). All scores are reported as accuracy (\%). Average denotes the overall accuracy over all MCQ items. \textbf{Bold} and \underline{underlined} values indicate the best and the second-best performance among all open-source and domain-specific models. By leveraging just 10\% to 30\% of AGBASE~\cite{agmmu}, AgroNVILA (SFT) demonstrates robust adaptation.}
% achieving exceptional performance in "Symptom" recognition (89.9\%), significantly surpassing even proprietary models like GPT-o4-mini.}
\label{tab:agmmu_mcq_detailed}
\renewcommand{\arraystretch}{1.1}
\resizebox{\linewidth}{!}{
\begin{tabular}{l c ccccc c}
\toprule
\textbf{Model} & \textbf{Size} & \textbf{Disease} & \textbf{Insect/Pest} & \textbf{Species} & \textbf{Management} & \textbf{Symptom} & \textbf{Average} \\
\midrule
\multicolumn{8}{l}{\textit{Proprietary Models}} \\
GPT-o4-mini~\cite{o4mini}       & -   & 77.9 & 85.4 & 90.3 & 93.8 & 84.3 & 86.5 \\
Gemini 1.5 Pro~\cite{gemini1.5}    & -   & 76.2 & 81.1 & 82.8 & 88.1 & 76.9 & 82.4 \\
Claude 3 Haiku    & -   & 62.1 & 71.2 & 52.8 & 81.5 & 52.0 & 63.8 \\
\midrule
\multicolumn{8}{l}{\textit{SOTA Open-sourced Models}} \\
LLaVA-1.5~\cite{llava1.5}        & 7B & 61.9 & 64.6 & 67.6 & 77.3 & 71.5 & 69.0 \\
Cambrian~\cite{cambrian}         & 8B  & 65.0 & 70.1 & 59.3 & 79.1 & \underline{86.0} & 72.8 \\
LLaVA-OneVision~\cite{llava_onevision}  & 8B  & 65.7 & 72.9 & \textbf{71.2} & \underline{85.9} & 78.8 & 75.4 \\
Qwen-VL~\cite{qwenvl} & 7B & 55.5 & 61.0 & 60.4 & 74.7 & 64.3 & 63.5 \\
NVILA-Lite-8B~\cite{nvila}             & 8B  & \textbf{68.6} & 70.8 & \underline{71.0} & 84.1 & 79.9 & 75.4 \\
\midrule
\multicolumn{8}{l}{\textit{Domain-Specific Models (Ours)}} \\
AgroNVILA (Zero-shot) & 8B  & 61.4 & 68.1 & 65.5 & 79.8 & 65.4 & 68.2 \\

AgroNVILA (10\% Data) & 8B & 64.3 & \underline{76.4} & 66.9 & 85.9 & 85.5 & \underline{76.5} \\

\textbf{AgroNVILA (30\% Data)} & \textbf{8B} & \underline{67.1} & \textbf{77.8} & 69.0 & \textbf{87.7} & \textbf{86.6} & \textbf{78.3} \\
\bottomrule
\end{tabular}
}
% \vspace{-2em}
\end{table*}

\subsection{Further Discussion} 

\begin{figure*}[!t]
  \centering
  
  % --- Subfigure (a): View Performance ---
  \begin{subfigure}{0.45\linewidth}
    \centering
    % 请将路径替换为你刚才用 Python 生成的图片的实际路径
    \includegraphics[width=\linewidth]{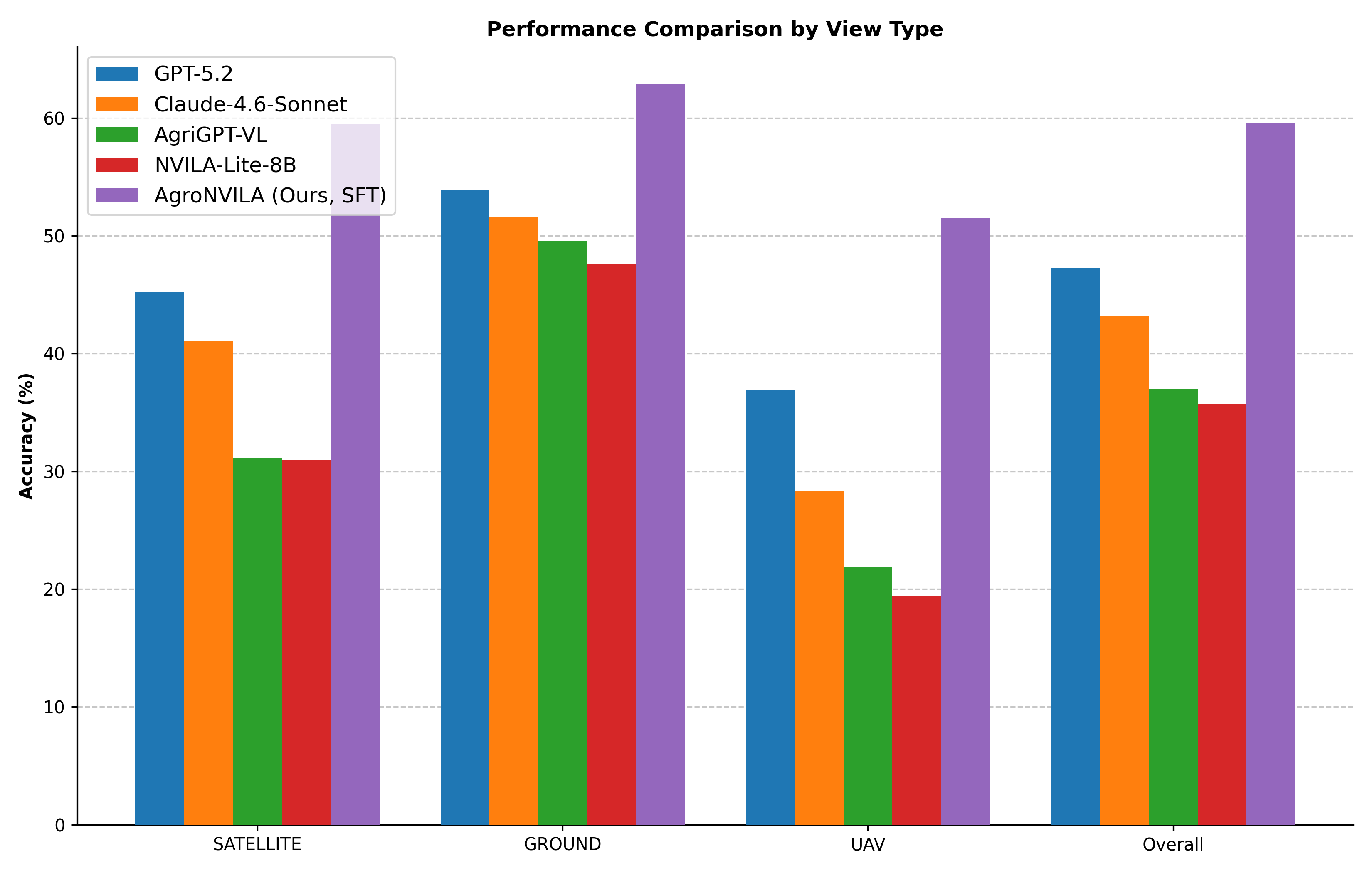}
    \caption{Performance comparison by view type}
    \label{fig:view_performance}
  \end{subfigure}
  % \hfill % 自动填充中间的空白间距，使两张图贴紧两边对齐
  % --- Subfigure (b): Data Scaling ---
  \begin{subfigure}{0.54\linewidth}
    \centering
    % 请将路径替换为你刚才用 Python 生成的图片的实际路径
    \includegraphics[width=\linewidth]{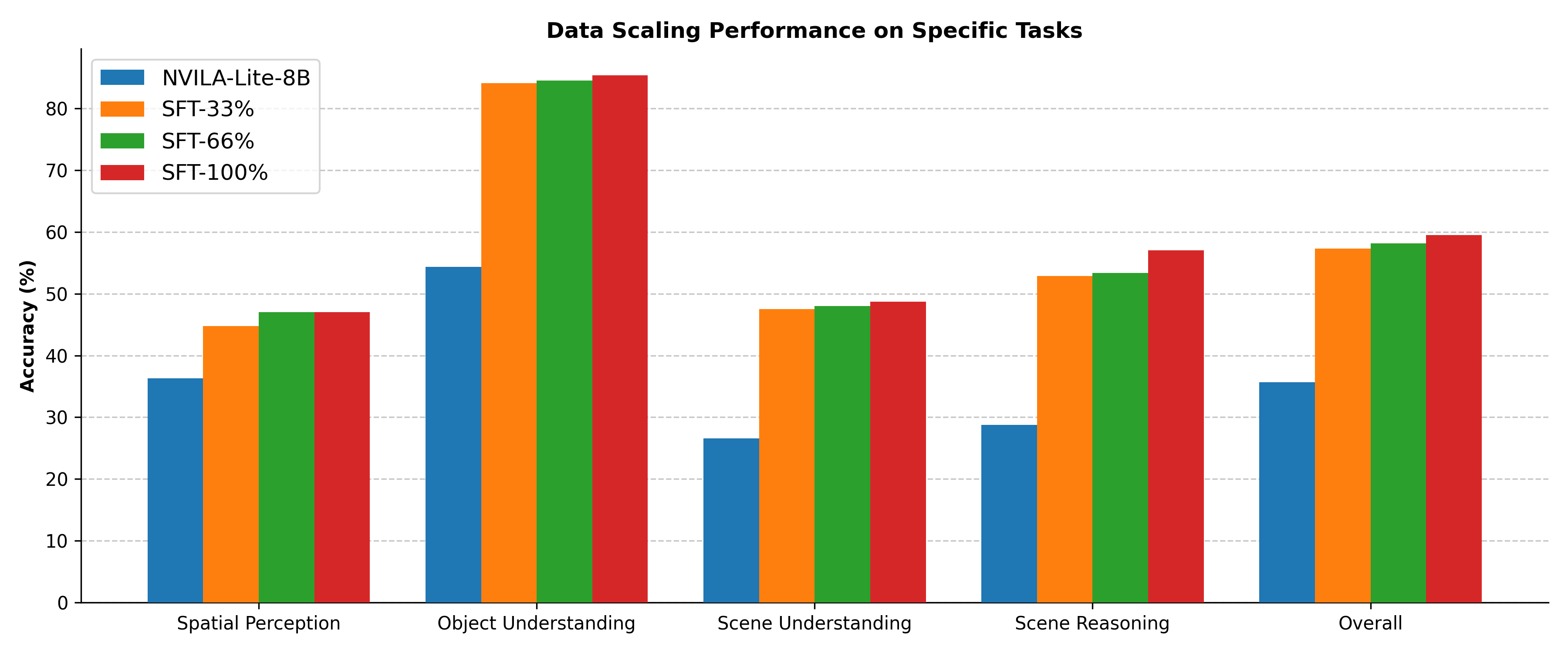}
    \caption{Data scaling on coarse-grained cognitive tasks}
    \label{fig:scaling_performance}
  \end{subfigure}
  
  \caption{\textbf{Comprehensive Diagnostic Analysis on AgroMind.} \textbf{(a)} Evaluation across different spatial perspectives demonstrates the effective mitigation of Ground-level bias. \textbf{(b)} The scaling trajectory exposes a strict representational plateau, where macro-topological tasks saturate rapidly despite increased data volume.}
  \label{fig:combined_analysis}
  \vspace{-1.5em}
\end{figure*}

\textbf{View Type Analysis.}
As Fig.~\ref{fig:view_performance} illustrates, all baseline models exhibit a severe \textit{Ground-level bias}, heavily underperforming on macro-scale views (e.g., Claude-4.6-Sonnet drops from 51.61\% on Ground to 28.31\% on UAV). This confirms that generic vision encoders lack the macroscopic topological priors required to decode top-down agricultural scenes. Crucially, fine-tuning on the AgroOmni corpus effectively mitigates this architectural blind spot, evidenced by highly asymmetric performance gains. Full-corpus SFT yields a moderate improvement on the already-familiar Ground view (+15.31\%), but triggers massive accuracy surges of \textbf{+32.12\%} (19.40\% $\rightarrow$ 51.52\%) and \textbf{+28.53\%} (30.97\% $\rightarrow$ 59.50\%) on UAV and Satellite views, respectively. This disproportionate growth empirically proves that AgroOmni provides the essential spatial signals necessary to rectify the perspective distortion inherent in current MLLMs.

% \begin{table}[!t]
% \centering
% \tiny
% \caption{Performance comparison by view type on the AgroMind benchmark. We report Accuracy (\%) for each view type, overall micro-average.}
% \label{tab:agromind_view_results}
% \resizebox{0.8\linewidth}{!}{\begin{tabular}{>{\centering\arraybackslash}m{2cm}cccc}
% \toprule
% Model & SATELLITE & GROUND & UAV & Overall \\
% \midrule
% GPT-5.2 & \underline{45.23} & \underline{53.85} & \underline{36.94} & \underline{47.29} \\
% Claude-4.6-Sonnet & 41.07 & 51.61 & 28.31 & 43.15 \\
% \midrule
% AgriGPT-VL & 31.13 & 49.58 & 21.90 & 36.97 \\
% NVILA-Lite-8B & 30.97 & 47.60 & 19.40 & 35.68 \\
% \midrule
% % SFT-33\% & 55.66 & 61.63 & 57.60 & 58.38 \\
% % SFT-66\% & \underline{58.18} & \underline{62.59} & \underline{59.17} & \underline{60.12} \\
% SFT-100\% & \textbf{59.39} & \textbf{64.21} & \textbf{61.10} & \textbf{61.61} \\
% \bottomrule
% \end{tabular}
% }
% \end{table}

\textbf{Data Scaling Analysis.}
To systematically diagnose cognitive bottlenecks, we analyze the scaling trajectory across four coarse-grained task categories (Fig. \ref{fig:scaling_performance}). The results reveal that while Object and Scene categories experience rapid semantic alignment, Spatial Perception exhibits a persistent structural lag. Specifically, exposing the model to merely 33\% of AgroOmni triggers a massive overall gain of +21.67\% (35.68\% $\rightarrow$ 57.35\%), largely driven by the explosive growth in \textit{Object Understanding} (54.32\% $\rightarrow$ 84.12\%). However, the improvement in \textit{Spatial Perception} remains significantly constrained, yielding only a marginal 8.42\% (36.31\% $\rightarrow$ 44.73\%). And when scaling the training ratio from 33\% to 100\%, all four tasks yield only marginal performance gains compared with the 33\% baseline. The results reveal an inherent architectural ceiling that while SFT can effectively teach the model what it perceives, it struggles to remedy the fundamental absence of macro-level spatial priors inherent in frozen vision encoders. Thus, although AgroOmni provides a high-density learning signal, the enduring gap in spatial tasks underscores the urgent need for future research to address spatial-centric architectural alignment in the agricultural domain.

\begin{figure}[t]
  \centering
  \includegraphics[width=\textwidth]{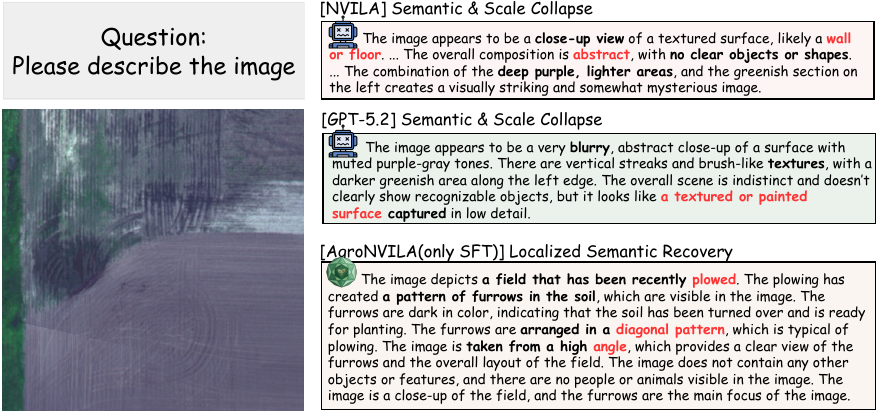}
  \caption{Scale Confusion and Semantic Collapse Case. Compared to current MLLMs, our AgroOmni dataset corrects the scale and perspective biases of generic MLLMs, enabling accurate agronomic interpretation of macro-scale agricultural scenes.}
  \label{fig:ablationcase}
  % \vspace{-1.5em}
\end{figure}

\textbf{Case Study.}
Fig.~\ref{fig:ablationcase} visualizes the representational deficits of current MLLMs using a minimalist prompt to probe unconditioned visual priors. Confronted with a macro-scale agricultural landscape, generic models (e.g., GPT-5.2, NVILA) suffer catastrophic scale confusion and semantic collapse. Driven by their inherent Ground-level bias, they erroneously decode the top-down topology as a micro-scale ``close-up view'' of a ``wall'' or ``painted surface.'' Conversely, our AgroNVILA (SFT), aligned via the AgroOmni corpus, demonstrates robust semantic recovery. It completely discards the abstract illusion and accurately grounds the visual features to expert agronomic semantics (identifying a ``plowed or tilled section of the field''). This stark qualitative contrast empirically validates AgroOmni's crucial role in rectifying the severe perspective distortions inherent in foundational vision encoders.

\section{Conclusion}

Agricultural multimodal reasoning faces challenges from scale and perspective ambiguities, as existing models often focus on \textit{Ground-level} data, leading to confusion and logic drift in multi-altitude agricultural tasks. To address this, we introduced \textbf{AgroOmni}, a large-scale, multi-view VQA training dataset that integrates Ground, UAV, and Satellite views, consisting of 268,119 QA pairs covering 56 specialized task categories across 14 task types, capturing diverse spatial topologies and scales for comprehensive reasoning. 
Building on the AgroOmni dataset, we proposed \textbf{AgroNVILA}, effectively bridging the perceptual gaps between disparate altitudes and outperforming proprietary SOTA models like GPT-5.2 (47.29\%) with a new state-of-the-art (59.86\%) on AgroMind. Diagnostic evaluations on AgMMU further reveal that spatial scales from Ground to Satellite are heterogeneous and show AgroNVILA’s rapid adaptation with minimal data fine-tuning, proving that AgroOmni provides a high-density learning corpus, enabling models to bridge the macro-to-micro gap.
% Our experiments on the AgroMind benchmark demonstrate that AgroNVILA achieves a new state-of-the-art performance (62.47\%), significantly outperforming the second-best model GPT-5.2 (47.29\%). 
These results underline the effectiveness of the AgroOmni dataset and the AgroNVILA model in advancing multi-view agricultural reasoning. Future works will focus on enhancing model scalability and deploying AgroNVILA in real-world agricultural systems.

\section*{Acknowledgments and Disclosures}
We thank the reviewers for their constructive comments. This work was supported in part by the National Natural Science Foundation of China under Grant T2125006 and Grant 42401415; in part by Shenzhen Science and Technology Program under Grant KCXFZ20240903093759004 and Grant KJZD20230923115106012. Yang Liu is supported in part by
the Postdoctoral Fellowship Scheme of The Chinese University of Hong Kong.

\textbf{Competing interests.} The authors declare no competing financial interests.

% \section*{References}

% References follow the acknowledgments in the camera-ready paper. Use unnumbered first-level heading for
% the references. Any choice of citation style is acceptable as long as you are
% consistent. It is permissible to reduce the font size to \verb+small+ (9 point)
% when listing the references.
% Note that the Reference section does not count towards the page limit.
% \medskip

{
%\small
\bibliographystyle{IEEEtran}
\bibliography{reference}
}

%%%%%%%%%%%%%%%%%%%%%%%%%%%%%%%%%%%%%%%%%%%%%%%%%%%%%%%%%%%%
\clearpage

\appendix
% % ---------------------------------------------------------------
% % TODO REVIEW: Replace with your title
% \title{\includegraphics[width=0.05\linewidth]{Figure/AgroNVILA.png}AgroNVILA: Perception-Reasoning Decoupling \\for Multi-view Agricultural Multimodal \\Large Language Models\\Supplementary Material} 

% % TODO REVIEW: If the paper title is too long for the running head, you can set
% % an abbreviated paper title here. If not, comment out.
% % \titlerunning{Abbreviated paper title}

% % TODO FINAL: Replace with your author list. 
% Second Author\inst{2,3}\orcidlink{1111-2222-3333-4444} \and
% Third Author\inst{3}\orcidlink{2222--3333-4444-5555}}

% % TODO FINAL: Replace with an abbreviated list of authors.

% % TODO FINAL: Replace with your institution list.
% Springer Heidelberg, Tiergartenstr.~17, 69121 Heidelberg, Germany

% \maketitle
% \vspace*{1cm} % 顶部留白
\begin{center}
    % 主标题：使用 \Large 完全等效于 llncs 的正文主标题字号
    {\Large \bfseries \includegraphics[width=0.05\linewidth]{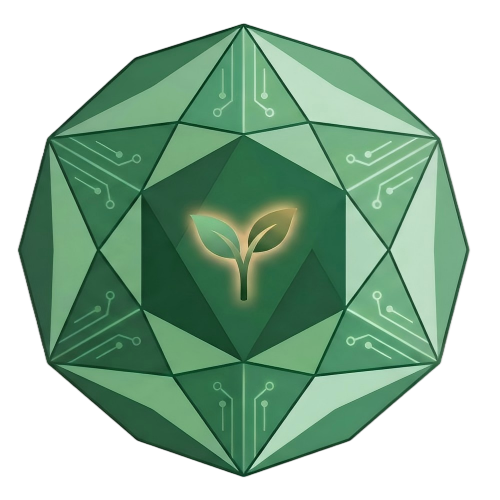} AgroOmni: A Large-Scale Multi-view Agricultural Dataset for Cross-Scale Multimodal Reasoning \par}
    
    \vspace{0.4cm} % 主副标题间距
    
    % 副标题：使用 \large 标明附录身份
    {\large \bfseries Supplementary Material \par}
\end{center}
\vspace{0.8cm} % 底部留白，拉开与正文的距离

% % \begin{abstract}
% %   The abstract should concisely summarize the contents of the paper. 
% %   While there is no fixed length restriction for the abstract, it is recommended to limit your abstract to approximately 150 words.
% %   Please include keywords as in the example below. 
% %   This is required for papers in LNCS proceedings.
% %   \keywords{First keyword \and Second keyword \and Third keyword}
% % \end{abstract}

% % 目录
% \renewcommand{\thesection}{\arabic{section}} 
% \setcounter{section}{0} % 从1开始计数

% % 2. 引擎层面的修改：解决“跳转回正文”的逻辑冲突
% % 给附录的跳转 ID 加上 "A" 前缀，使其与正文的 section.4 区分开
% \renewcommand{\theHsection}{A.\thesection}

% % 3. (可选) 引用层面的修改：让 ~\ref 自动显示为 "Appendix 4" 而不是 "Section 4"
% ~\refalias{section}{appendix}

\section*{Table of Contents}

\begin{enumerate}[label=\Alph*., leftmargin=*]
    \item \textbf{\hyperref[suppl:method]{More Detail about RL}}
    \begin{enumerate}[label=\Alph{enumi}.\arabic*., leftmargin=*]
        \item Extended Details on Reinforcement Learning Alignment
        \item Detailed Reward Formulations
        \item KL-aware Regularization and Advantage Dampening
    \end{enumerate}

    \item \textbf{\hyperref[suppl:reproducibility]{Reproducibility and Implementation Details}}
    \begin{enumerate}[label=\Alph{enumi}.\arabic*., leftmargin=*]
        \item Hardware and Environment
        \item Stage-wise Training Pipeline
    \end{enumerate}

    \item \textbf{\hyperref[suppl:ablations]{View Ablation and Seed Stability}}

    \item \textbf{\hyperref[suppl:general_retention]{General Capability Retention}}
    
    \item \textbf{\hyperref[suppl:dataset]{Extended Details of the AgroOmni Dataset}}
    \begin{enumerate}[label=\Alph{enumi}.\arabic*., leftmargin=*]
        \item Data Collection and Multi-view Composition
        \item Data Pre-processing
        \item Question Generation
        \item Processing for Each Dataset
    \end{enumerate}

    \item \textbf{\hyperref[suppl:agromind_eval_details]{Model Details of AgroMind Evaluation}}

    \item \textbf{\hyperref[suppl:case_study]{Qualitative Analysis and Case Studies}}

    \item \textbf{\hyperref[suppl:bi]{Broader Impacts}}

    \item \textbf{\hyperref[suppl:laf]{Limitations and Future Work}}

    \item \textbf{\hyperref[suppl:ec]{Ethical Considerations}}
\end{enumerate}

\section{More Detail about RL}
\label{suppl:method}

\subsection{Extended Details on Reinforcement Learning Alignment}
\label{suppl:RL-detail}

\textbf{PPO.} Proximal Policy Optimization (PPO)~\cite{PPO} is a foundational actor-critic reinforcement learning algorithm widely adopted for model alignment. It maintains two distinct networks: a policy model $\pi_\theta$ and an auxiliary value model $V_\omega$ that estimates the baseline for the advantage function $\hat{A}^{\text{PPO}}$. Given a prompt $q$ and a generated response $o_i$, the policy is optimized by maximizing a clipped surrogate objective:
\begin{equation}
J_{\text{PPO}}(\theta) = \mathbb{E}_{q, o} \left[ \min\!\left(\phi_i \hat{A}^{\text{PPO}}_i,\, \text{clip}(\phi_i, 1\!-\!\varepsilon, 1\!+\!\varepsilon) \hat{A}^{\text{PPO}}_i\right) - \beta D_{\text{KL}}(\pi_\theta \| \pi_{\text{ref}}) \right],
\end{equation}
where $\phi_i = \frac{\pi_\theta(o_i|q)}{\pi_{\text{old}}(o_i|q)}$ denotes the probability ratio between the current and previous policies, $\varepsilon$ is the clipping hyperparameter, and $\beta$ controls the Kullback-Leibler (KL) divergence penalty against a reference model $\pi_{\text{ref}}$. 

\textbf{GRPO.} Group Relative Policy Optimization (GRPO)~\cite{GRPO} optimizes policies by removing the auxiliary critic network and computing advantages directly within a group of $G$ sampled responses $\{o_1, \dots, o_G\}$ for a given prompt $q$. The normalized advantage is:
\begin{equation}
\hat{A}^{\text{GRPO}}_i = \frac{r_i - \mu_G}{\sigma_G + \varepsilon},
\end{equation}
where $\mu_G$ and $\sigma_G$ are the group-wise mean and standard deviation. The policy $\pi_\theta$ is updated by maximizing the clipped surrogate objective with a KL penalty:
\begin{equation}
J_{\text{GRPO}}(\theta) = \mathbb{E} \left[ \frac{1}{G} \sum_{i=1}^G \min\!\left(\phi_i \hat{A}^{\text{GRPO}}_i,\, \text{clip}(\phi_i, 1\!-\!\varepsilon, 1\!+\!\varepsilon) \hat{A}^{\text{GRPO}}_i\right) - \beta D_{\text{KL}}(\pi_\theta \| \pi_{\text{ref}}) \right].
\end{equation}
% Despite its computational efficiency, GRPO lacks a mechanism to differentiate between heterogeneous tasks. In benchmarks like AgroMind, easy high-frequency tasks (e.g., binary judgment) dominate gradient updates, providing insufficient learning signals for harder reasoning tasks such as spatial localization.

\textbf{DRPO.} Domain-aware Relative Policy Optimization (DRPO)~\cite{DRPO} addresses the performance imbalance caused by skewed data distributions and varying task difficulties. To prevent abundant, low-difficulty prompts from overwhelming the optimization process, DRPO introduces an adaptive, hierarchical reward scaling mechanism. Prompts are first categorized by their respective domains $d$, and their per-prompt reward vectors are clustered (e.g., via K-means) to identify intra-domain difficulty strata $c$. A scaling temperature $T_{d,c}$ is then computed for each cluster as a function of its sample count $N_{d,c}$ (domain rarity) and average reward $\mu_{d,c}$ (task difficulty). Samples from common, easy tasks are assigned a higher temperature (resulting in a smaller gradient weight), whereas rare or difficult samples are upweighted. The advantage in DRPO is derived by scaling the standard group-relative advantage:
\begin{equation}
\hat{A}^{\text{DRPO}}_i = \frac{1}{T_{d,c}} \hat{A}^{\text{GRPO}}_i.
\end{equation}
These scaled advantages are subsequently renormalized across the training batch before updating the policy using the standard surrogate objective $J_{\text{GRPO}}(\theta)$. By dynamically modulating the gradient signals without requiring an auxiliary critic, DRPO ensures balanced learning across heterogeneous modalities and protects the learning signals of complex reasoning tasks.

\subsection{Detailed Reward Formulations}
\label{suppl:appendix_reward}

During the reinforcement learning alignment phase, we employ a combination of one main accuracy reward and two lightweight auxiliary rewards that jointly capture task correctness, output validity, and spatial consistency in agricultural scenes. Table~\ref{tab:suppl_reward_design} summarizes the reward components and the task-aware scoring for $r_{\text{task}}$.

\textbf{Accuracy reward.}
The primary objective is task accuracy, for which we compute a task-aware reward $r_{\text{task}}(o, q)$ per question type. We use exact letter matching for single-choice questions, a continuous relative error score for counting, the geometric mean of set-IoU and recall for multi-choice selection, IoU-based scoring for bounding box prediction, and ROUGE-L with a length penalty for open-ended answers. For short-answer categories such as growth-stage classification and farmland description, we further apply ordinal-distance scoring and triplet-level matching to grant partial credit to near-miss predictions.

\textbf{Spatial consistency reward.}
For spatial tasks (e.g., boundary detection and localization), we additionally encourage geometric faithfulness via a spatial reward $r_{\text{spatial}}(o, q)$. Given a predicted box $b$ and ground-truth region $S$, we define $r_{\text{spatial}}$ as the intersection-over-union between $b$ and $S$, which nudges the policy toward accurate delineation of cultivated areas and salient anomalies.

\textbf{Auxiliary rewards.}
We also include a lightweight format reward $r_{\text{fmt}}(o, q)$ that checks whether the output conforms to the expected syntax (e.g., valid option letters, numeric counts, or coordinate tuples). This term is inexpensive to compute and helps stabilize RL training by discouraging degenerate or unparsable responses.

\textbf{Combined reward.}
The final scalar reward supplied to the policy optimization objective is a weighted combination:
\begin{equation}
r = \lambda_{\text{task}} r_{\text{task}} + \lambda_{\text{spatial}} r_{\text{spatial}} + \lambda_{\text{fmt}} r_{\text{fmt}}.
\end{equation}
In our experiments, we set $(\lambda_{\text{task}}, \lambda_{\text{spatial}}, \lambda_{\text{fmt}}) = (0.8, 0.1, 0.1)$, which balances task accuracy with spatial grounding and output validity while keeping the reward computation computationally efficient.

\begin{table}[t]
  \centering
  \caption{\textbf{Task-specific reward functions.}}
  \label{tab:suppl_reward_design}
  %\scriptsize
  \begin{tabular}{@{}ll@{}}
    \toprule
    \textbf{Task} & \textbf{Reward Design Details} \\
    \midrule
    Single-choice / VQA & Reward = 1.0 for exact match, 0 otherwise. \\
    Counting & Reward = $\max(0, 1 - |\text{Ans}-\text{GT}|/\max(|\text{GT}|,1))$. \\
    Multi-choice & Reward = geometric mean of set-IoU and recall. \\
    Visual grounding / BBox & Reward = IoU(pred, GT). \\
    Open-ended & Reward = ROUGE-L with length penalty. \\
    Short-answer (e.g., growth-stage) & Reward = ordinal-distance and triplet-level matching. \\
    Boundary detection \& localization & Reward = IoU(pred, GT). \\
    Format (auxiliary) & Syntax check (option letters, counts, coordinates). \\
    \bottomrule
  \end{tabular}
\end{table}

\subsection{KL-aware Regularization and Advantage Dampening}
\label{suppl:appendix_kl}

The hierarchical scaling mechanism utilized in DRPO, while effective for balancing heterogeneous tasks, can significantly increase the variance of normalized advantages. This may lead to optimization instability, where a small subset of responses with extreme advantages dominates the gradient updates. To mitigate this, we implement a \textbf{KL-aware regularization} strategy, following the approach in QoQ-Med \cite{DRPO}, to dampen advantages that exhibit excessive policy drift.

Specifically, we control the optimization process through two complementary mechanisms:

\textbf{Global KL Penalty:} We incorporate a standard scalar KL penalty $\beta\, D_{\mathrm{KL}}(\pi_\theta \| \pi_{\mathrm{ref}})$ into the surrogate objective. We set $\beta=0.05$ to ensure the learned policy remains within a trust region of the reference model, thereby preventing catastrophic forgetting of the foundational visual-semantic priors acquired during the SFT stage.

\textbf{Inverse-Linear Advantage Regularizer:} To further stabilize training against outliers, we apply an inverse-linear dampening factor to the scaled advantages based on their question-level KL divergence. For each response $o_{(q,i,t)}$, we first compute the KL divergence:
\begin{equation}
k_{(q,t)} = \sum_{j=1}^{L} \left[ \pi_\theta(o_j|q) \log \frac{\pi_\theta(o_j|q)}{\pi_{\mathrm{ref}}(o_j|q)} \right],
\end{equation}
where $L$ is the sequence length. We then define a multiplicative dampening factor $m_{(q,i,t)}$ to adjust the scaled advantage prior to batch-wise renormalization:
\begin{equation}
m_{(q,i,t)} = \frac{t_p}{t_p + \max\big( s^{\mathrm{scaled}}_{(q,i,t)} \cdot k_{(q,t)},\, 0 \big)},
\end{equation}
where $s^{\mathrm{scaled}}_{(q,i,t)}$ denotes the advantage after domain and cluster temperature scaling, and $t_p$ is the $p$-th percentile of the product $\{ s^{\mathrm{scaled}} \cdot k \}$ within the mini-batch (we set $p=0.9$). This regularization design offers \textbf{several key advantages:}
\begin{itemize}
    \item \textbf{Deterrence of Shortcuts:} It selectively dampens responses that achieve high rewards by deviating excessively from the reference model. In agricultural reasoning, such samples often represent overconfident ``statistical shortcuts'' rather than robust agronomic logic.
    \item \textbf{Computational Efficiency:} Since $k_{(q,t)}$ is already required for the primary KL penalty in the loss function, the dampening factor $m_{(q,i,t)}$ is computed with negligible overhead, requiring no additional forward passes.
    \item \textbf{Stability:} By suppressing high-variance outliers, this mechanism ensures smoother policy updates even under the extreme task heterogeneity of the AgroMind benchmark.
\end{itemize}

\section{Reproducibility and Implementation Details}
\label{suppl:reproducibility}

To facilitate full reproducibility of our experimental baselines and benchmarking results on the AgroOmni dataset, we detail the hardware environment and the stage-wise training hyperparameters below.

\subsection{Hardware and Environment}
All experiments, including both supervised fine-tuning and policy optimization, are conducted on a computing cluster equipped with NVIDIA A100 (80GB) GPUs. Our implementation is built upon the official NVILA codebase. We extend its core training pipeline to robustly support advanced reinforcement learning alignment frameworks, specifically integrating multi-rollout generation and advantage computation for methods like GRPO and DRPO. The full experimental campaign reported in this paper used about 370 GPU-hours on these devices.

\subsection{Stage-wise Training Pipeline}
The training pipeline for establishing our robust multimodal baselines is executed in a rigorous two-stage progression. The comprehensive hyperparameters for both stages are summarized in Table~\ref{tab:hyperparameters}.

\textbf{Stage 1: Supervised Fine-Tuning (SFT).} In this initial alignment phase, we fine-tune the foundational multimodal framework using the complete AgroOmni dataset. To ensure parameter efficiency while effectively acquiring domain-specific spatial knowledge, we freeze the language backbone and inject Low-Rank Adaptation (LoRA) modules ($r=128, \alpha=256$) into the base LLM. The vision encoder remains strictly frozen, while the vision-language projector is fully updated to map visual tokens into the LLM's continuous latent space. The model is trained for a single epoch with a maximum sequence length of 5120.

\textbf{Stage 2: Reinforcement Learning Alignment.} In the subsequent reinforcement learning stage, we unfreeze the LLM backbone for full-parameter training alongside the vision-language projector, while the vision encoder remains strictly frozen. To align the model's decision-making with expert agronomic logic, we employ group-based policy optimization (GRPO and DRPO). During training, we process each prompt by sampling a group of $G=8$ rollouts using a sampling temperature of $0.9$ and a maximum generation length of $128$ tokens. Optimization is driven by AdamW with a constant learning rate of $2 \times 10^{-7}$ for the LLM (scaled by $0.1$ for the projector), a weight decay of $0.01$, and PPO-style clipping ($\varepsilon=0.2$). Notably, when applying DRPO, the objective is further regularized by a KL penalty ($\beta=0.05$) anchored to the reference policy. To manage the intensive memory footprint, we enable gradient checkpointing and maintain a global batch size of 32, configured via a per-device batch size of 1 prompt and 4 gradient accumulation steps. The reward computation incorporates a format-checking term (weight $0.1$), and samples yielding near-uniform group rewards are dynamically skipped (threshold $0.05$) to prevent low-signal updates.

\begin{table}[h]
\centering
\caption{\textbf{Global Hyperparameters for the two-stage training pipeline.}}
\label{tab:hyperparameters}
\resizebox{0.95\linewidth}{!}{
\begin{tabular}{@{}lcc@{}}
\toprule
\textbf{Hyperparameter} & \textbf{Stage 1: SFT} & \textbf{Stage 2: GRPO \& DRPO} \\
\midrule
Trainable Modules & LLM (LoRA), Projector & LLM (full), Projector \\
Frozen & Vision Encoder & Vision Encoder \\
LoRA Config ($r, \alpha, \text{dropout}$) & $128, 256, 0.05$ & -- \\
Global Batch Size & 32 & 32 \\
Per-Device Batch Size (prompts) & 4 & 1 \\
Group Size $G$ (rollouts per prompt) & -- & 8 \\
Gradient Accumulation & 1 & 4 \\
Learning Rate (LR) & $2 \times 10^{-5}$ & $2 \times 10^{-7}$ \\
Projector LR Scale & -- & $0.1$ \\
LR Schedule & Cosine Decay & Constant \\
Warmup Ratio & 0.03 & 0.0 \\
Weight Decay & 0.0 & 0.01 \\
Max Sequence Length & 5120 & 5120 \\
Max New Tokens (generation) & -- & 128 \\
Temperature (sampling in DRPO) & -- & 0.9 \\
KL Coefficient $\beta$ (Only DRPO) & -- & 0.05 \\
Clip Range $\varepsilon$ & -- & 0.2 \\
Reward Format Weight & -- & 0.1 \\
Optimizer & AdamW & AdamW (w/ KL Penalty) \\
Epochs & 1 & 1 \\
Precision & bfloat16 & bfloat16 \\
\bottomrule
\end{tabular}
}
\end{table}

\section{View Ablation and Seed Stability}
\label{suppl:ablations}

We separate the effect of view composition from data volume with two controlled settings. Both settings use NVILA-Lite-8B, the SFT recipe of Table~\ref{tab:hyperparameters}, and seed 42. All conditions within one setting share the same sample budget and the same number of updates.

\textbf{Full-budget comparison.} The first setting uses 192,800 samples and 6,025 updates. The Ground-only condition uses every Ground sample and fills the remaining budget by stratified resampling of the same view. The multi-view condition keeps the natural view ratio of AgroOmni. Table~\ref{tab:ablation_full} reports the result.

Ground-only training keeps the Ground score (63.22\%) but transfers poorly, with 35.96\% on Satellite and 23.02\% on UAV. The multi-view condition reaches 58.33\% overall and gains 22.36 points on Satellite and 28.04 points on UAV. More Ground data therefore cannot replace direct coverage of the other views.

\begin{center}
\centering
\small
\captionof{table}{\textbf{Full-budget view ablation.} Both conditions use 192,800 samples and 6,025 updates. \textbf{Bold} marks the best value in each column.}
\label{tab:ablation_full}
\begin{tabular}{lcccc}
\toprule
Training data & Satellite & Ground & UAV & Overall \\
\midrule
Ground-only & 35.96 & \textbf{63.22} & 23.02 & 44.71 \\
Proportional multi-view & \textbf{58.32} & 61.40 & \textbf{51.06} & \textbf{58.33} \\
\midrule
Change & +22.36 & $-$1.82 & +28.04 & +13.62 \\
\bottomrule
\end{tabular}
\end{center}

\textbf{Matched-size comparison.} The second setting fixes every condition at 18,528 samples and 579 updates. The budget follows the UAV pool, rounded down to a multiple of the global batch size. Table~\ref{tab:ablation_matched} compares four conditions.

Each single-view model is strongest on its own view and transfers poorly to unseen views. The multi-view condition reaches 51.82\% overall. It exceeds the best single-view condition (Satellite-only, 44.56\%) by 7.26 points and Ground-only (40.24\%) by 11.58 points.

\begin{center}
\centering
\small
\captionof{table}{\textbf{Matched-size view ablation.} Every condition uses 18,528 samples and 579 updates. \textbf{Bold} marks the best value in each column.}
\label{tab:ablation_matched}
\begin{tabular}{lcccc}
\toprule
Training data & Satellite & Ground & UAV & Overall \\
\midrule
Ground-only & 31.50 & \textbf{59.13} & 17.65 & 40.24 \\
Satellite-only & \textbf{53.86} & 46.24 & 16.81 & 44.56 \\
UAV-only & 30.31 & 45.75 & \textbf{54.52} & 40.58 \\
Proportional multi-view & 48.17 & 57.85 & 46.77 & \textbf{51.82} \\
\bottomrule
\end{tabular}
\end{center}

\textbf{Seed stability.} We repeat the full-corpus SFT run with seeds 42, 43, and 44. Table~\ref{tab:seed_stability} lists the individual runs and the mean. The overall score is 59.53 $\pm$ 0.03\%, which shows that the recipe is insensitive to the random seed.

\begin{center}
\centering
\small
\captionof{table}{\textbf{Seed stability of the full-corpus SFT run.} All three runs use the complete AgroOmni corpus and the SFT recipe of Table~\ref{tab:hyperparameters}.}
\label{tab:seed_stability}
\begin{tabular}{ccccc}
\toprule
Seed & Satellite & Ground & UAV & Overall \\
\midrule
42 & 59.50 & 62.91 & 51.52 & 59.52 \\
43 & 59.51 & 63.01 & 51.19 & 59.51 \\
44 & 59.33 & 63.05 & 51.88 & 59.56 \\
\midrule
Mean $\pm$ std & 59.45 $\pm$ 0.10 & 62.99 $\pm$ 0.07 & 51.53 $\pm$ 0.35 & 59.53 $\pm$ 0.03 \\
\bottomrule
\end{tabular}
\end{center}

\section{General Capability Retention}
\label{suppl:general_retention}

Domain-specific fine-tuning can erode the general visual ability of the base model. We therefore evaluate NVILA-Lite-8B and AgroNVILA on MMBench-DEV-EN, which has 1,164 scoring units. Both models use the same answer-extraction protocol.

Table~\ref{tab:mmbench} reports the result. AgroNVILA scores 89.18\% and the base model scores 91.75\%. The gap is 2.58 points.

The two models agree on 1,015 items. The base model answers 53 items that AgroNVILA misses, and AgroNVILA answers 23 items that the base model misses. The drop is small, so it does not show catastrophic forgetting. It does show a measurable cost of specialization.

\begin{center}
\centering
\small
\captionof{table}{\textbf{General capability retention on MMBench-DEV-EN.} Both models use the same 1,164 scoring units.}
\label{tab:mmbench}
\begin{tabular}{lc}
\toprule
Model & MMBench-DEV-EN \\
\midrule
NVILA-Lite-8B & 91.75 \\
AgroNVILA (SFT) & 89.18 \\
\bottomrule
\end{tabular}
\end{center}

\section{Extended Details of the AgroOmni Dataset}
\label{suppl:dataset}
\subsection{Data Collection and Multi-view Composition}

The curation of AgroOmni is fundamentally driven by the necessity for a comprehensive, scale-aware, and ecologically diverse training corpus. In this subsection, we detail our data collection paradigm. We elaborate on the strategic integration of multi-scale viewpoints, the extensive geographical and scenario diversity covered by our sources, and the physical rationale behind the dataset's native domain imbalance which directly motivates our architectural innovations.

\textbf{Multi-scale Viewpoint Integration.}
% 多视角的作用和组成分析
Unlike previous agricultural datasets confined to a single perspective, AgroOmni spatially spans three distinct observational scales (shown in Fig.~\ref{fig:data_statistics}(b)) by integrating 12 diverse open-source datasets with a proprietary parcel collection (detailed in Table~\ref{tab:dataset_summary}), comprehensively covering real-world scenarios in modern precision agriculture. This cross-scale visual instruction tuning fully activates and endows the model with superior practical capabilities in the agricultural domain:

\begin{itemize}
    \item \textbf{Ground-view:} Focuses on micro-scale crop phenotypes and pathological features. For instance, in datasets such as ACFR, IP102, and the 2018 AI Challenge, the model is required to capture minute lesions on leaves or identify specific insect species. The rich data from this perspective provides high-fidelity texture priors for the model to master fine-grained Object Classification (OC) and Pest/Disease Diagnosis (PDD).
    
    \item \textbf{UAV-view:} Fills the meso-scale gap between the micro and macro levels. By leveraging datasets like OilPalmUAV and PhenoBench, the scenarios are extended to plot-level plant density estimation, weed distribution analysis, and complex Visual Prompt Reasoning (VPR). This perspective trains the model to develop strong dense object counting capabilities and local spatial topology awareness.
    
    \item \textbf{Satellite-view:} Endows the model with macro-scale global planning capabilities. By incorporating OAM-TCD, CropHarvest, and internally collected farmland vector data, the trained model handles large-scale land coverage analysis, Climate Type Reasoning (CTR), and Boundary Detection (BD), directly serving regional-level agricultural resource scheduling.
\end{itemize}

\textbf{Scenario Diversity.}
%情景、地理的多样性，加入物候变化
Beyond multi-scale perspective coverage, AgroOmni exhibits exceptional scene generalization. Our data sources transcend isolated greenhouses or controlled experimental fields, spanning diverse global climate zones—such as tropical rainforests, temperate plains, and arid regions—and complex agricultural topographies. The dataset covers dozens of staple and cash crops (e.g., wheat, maize, oil palm, tomato) while faithfully preserving challenging environmental interferences, including variable lighting, cloud-induced occlusions, dense weed cohabitation, and irregular field boundaries. This cross-regional, cross-species, and cross-environmental richness significantly raises the complexity upper bound of multimodal reasoning and cultivates generalization in realistic agricultural settings. Furthermore, to transcend the limitations of static single-frame recognition, we introduce the Temporal Variation (TV) dimension. By integrating GWHD~\cite{David2021Wheat} for crop growth stage evolution and CLCD~\cite{liu2022CLCD} for bi-temporal farmland change detection, we equip the model with dynamic monitoring and longitudinal comparative reasoning capabilities. This temporal dimension aligns with the long-cycle nature of real-world agricultural production, addressing a critical deficiency in existing training corpora.

\textbf{Analysis of Real-world Domain Imbalance.}
%对数据集组成不均衡的分析
It is crucial to note that AgroOmni exhibits a pronounced native imbalance, mirroring the realistic acquisition costs and long-tail characteristics of agricultural data: Ground and Satellite modalities dominate the perspective distribution ( Fig.~\ref{fig:data_statistics}(b)), while foundational perception tasks (e.g., disease classification) significantly outweigh complex planning and temporal reasoning tasks ( Fig.~\ref{fig:data_statistics}(a)). We intentionally preserve this heterogeneous, "extremes-dominant" distribution rather than resorting to aggressive undersampling, as it constitutes the most authentic mapping of real-world agricultural intelligence. However, performing conventional SFT on such a skewed corpus inevitably drives models toward "perspective bias" and "statistical shortcuts"—overfitting to data-dense single perspectives while failing to internalize reasoning logic for complex long-tail scenarios. This distribution challenge highlights the inherent complexity of multi-scale agricultural data. It underscores the necessity for future models to go beyond generic visual encoders, requiring specialized spatial anchoring to resolve cross-view feature confusion. Furthermore, the task heterogeneity within AgroOmni suggests that standard learning strategies may require more adaptive, task-aware balancing to effectively internalize agronomic logic without over-fitting to dominant data categories.

\begin{table}[htbp]
    \centering
    \caption{\textbf{Detailed description of the datasets in AgroOmni.}} 
    \label{tab:dataset_summary}
    \renewcommand{\arraystretch}{0.9} % 压缩行距
    \setlength{\tabcolsep}{3pt}        % 缩小列内边距，减少换行
    \setlength{\aboverulesep}{0pt}     % 消除 \midrule 默认留白（修复断开的竖线）
    \setlength{\belowrulesep}{0pt}
    \renewcommand{\tabularxcolumn}[1]{m{#1}} % 强制要求 tabularx 的 X 列也使用垂直居中对齐
    \scriptsize
    \begin{tabularx}{\textwidth}{
        >{\centering\arraybackslash}m{1.8cm}   % 缩小 Dataset 列宽度以腾出 Year 列空间
        !{\vrule width 0.5pt}
        >{\centering\arraybackslash}m{0.8cm}   % Year
        !{\vrule width 0.5pt}
        >{\centering\arraybackslash}m{4.0cm} % Content
        !{\vrule width 0.5pt}
        >{\centering\arraybackslash}m{1.5cm}   % Resolution
        !{\vrule width 0.5pt}
        >{\raggedright\arraybackslash}X      % Link
    }
        \toprule
        \textbf{Dataset} & \textbf{Year} & \textbf{Content} & \textbf{Resolution} & \multicolumn{1}{c}{\textbf{Link}} \\
        \midrule
        ACFR Orchard Fruit Dataset \cite{bargoti2016deep} & 2016 & 
        pixel-level annotations for almond, apple, mango  & 
        308×202, 500×500 & 
        \url{https://data.acfr.usyd.edu.au/ag/treecrops/2016-multifruit/} \\
        \midrule
        
        2018 AI Challenge Dataset & 2018 & 
        Extensive crop leaf image collection featuring 3-level disease severity labels and 26 disease types across 9 crops. & 
        Variable & 
        \url{https://aistudio.baidu.com/datasetdetail/76075} \\
        \midrule

        IP102 Dataset \cite{ip102} & 2019 & 
        A large-scale agricultural pest dataset containing VOC-format bounding boxes for 102 insect species. & 
        Variable & 
        \url{https://github.com/xpwu95/IP102} \\
        \midrule

        Agriculture Vision Challenge \cite{Chiu_2020_CVPR} & 2020 & 
        Aerial multi-spectral (RGB + NIR) imagery equipped with pixel-level binary masks for detecting six common field anomalies. & 
        512×512 & 
        \url{https://github.com/SHI-Labs/Agriculture-Vision} \\
        \midrule
        
        OilPalmUAV Dataset \cite{zheng2021growing} & 2021 & 
        High-resolution UAV drone imagery providing instance-level bounding box annotations for categorizing five distinct growth states of oil palms. & 
        1024×1024 & 
        \url{https://github.com/rs-dl/MOPAD} \\
        \midrule

        CropHarvest Dataset \cite{tseng2021cropharvest} & 2021 & 
        Global remote sensing time-series combining satellite imagery and climate data with agricultural labels for multimodal crop classification. & 
        896×832, 960×896 & 
        \url{https://zenodo.org/records/5828893} \\
        \midrule

        GWHD Dataset \cite{David2021Wheat} & 2021 & Global wheat images and metadata equipped with bounding box annotations and four types of QA pairs for evaluating LMMs in object counting and visual reasoning. & 1024×1024 & \url{https://zenodo.org/records/5092309}
        \\
        \midrule

        CLCD Dataset \cite{liu2022CLCD} & 2022 & Bi-temporal Gaofen-2 satellite image pairs equipped with binary masks for cropland change detection. & 512×512 & \url{https://github.com/liumency/CropLand-CD} \\
        \midrule
        
        PhenoBench Dataset \cite{weyler2024pami}  & 2023 & 
        UAV-captured agricultural field images offering hierarchical annotations for crop/weed semantic segmentation and plant instance segmentation. & 
        1024×1024 & 
        \url{https://github.com/PRBonn/phenobench} \\
        \midrule

        Tomato Dataset \cite{zhang2025Tomato} & 2023 & Multi-illumination tomato images equipped with maturity labels, bounding boxes, and diverse VQA pairs for evaluating comparative crop maturity and global spatial reasoning. & 1280×720&\url{https://www.sciengine.com/CSD/doi/10.11922/11-6035.csd.2023.0154.zh} \\
        \midrule

        OAM-TCD Dataset \cite{veitchmichaelis2024oamtcdgloballydiversedataset} & 2024 & 
        Globally sourced aerial imagery paired with MS-COCO format bounding boxes and polygon masks for precise tree cover mapping. & 
        2048×2048 & 
        \url{https://zenodo.org/records/11617167} \\ 
        \midrule

        EarthVQA Dataset \cite{wang2024earthvqa} & 2024 & High-resolution remote sensing image-mask annotations equipped with 8-category pixel-level semantic masks for complex relational reasoning-based visual question answering. & 1024×1024& \url{https://github.com/Junjue-Wang/EarthVQA} \\
        \midrule

        Private Parcel Dataset & 2025 & 
        Internally collected multi-spectral imagery of cultivated parcels in several Chinese provinces, used for parcel coverage analysis. & 
        Variable & 
        Internal Collection \\
        \bottomrule
    \end{tabularx}
\end{table}

\begin{center}
\begin{minipage}{\linewidth}
\centering
\small
\captionof{table}{\textbf{Source coverage of AgroOmni.} The last column gives the number of QA pairs that each source contributes to the released corpus.}
\label{tab:coverage}
\begin{tabular}{@{}l l l l r@{}}
\toprule
Source & Region & Crops or targets & View & QA pairs \\
\midrule
ACFR Orchard Fruit & Australia & Almond, apple, mango & Ground & 15,520 \\
2018 AI Challenge & China & Crop leaves, 26 disease types & Ground & 64,641 \\
IP102 & Global & 102 insect pests & Ground & 84,579 \\
Agriculture-Vision & USA & Maize, soybean & Satellite & 15,142 \\
OilPalmUAV & Malaysia & Oil palm & UAV & 17,819 \\
CropHarvest & Global & Multiple crops & Satellite & 10,091 \\
GWHD & Global, nine countries & Wheat & Ground & 26,060 \\
CLCD & China & Cropland change & Satellite & 1,440 \\
PhenoBench & Germany & Sugar beet, maize, weed & UAV & 768 \\
Tomato & China & Tomato & Ground & 2,000 \\
OAM-TCD & Global & Tree cover & Satellite & 13,107 \\
EarthVQA & China & Land cover & Satellite & 4,352 \\
Rural & China, several provinces & Cultivated parcels & Satellite & 12,600 \\
\midrule
Total & & & & 268,119 \\
\bottomrule
\end{tabular}
\end{minipage}
\end{center}

\subsection{Data Pre-processing}
% 还要提一下AgroMind的去重
\label{suppl::preprocessing}

We adopt and extend the rigorous data processing pipeline established in AgroMind~\cite{li2025agromind}, implementing customized protocols to handle the heterogeneity of our data sources:
\begin{itemize}
\item \textbf{Manual Screening and Deduplication:}
A manual screening phase is conducted on augmented images to eliminate defective samples exhibiting artifacts, such as excessive content overlap or extreme flipping, which can otherwise distort semantic information and introduce noise.
\item \textbf{Format Conversion:} 
TIFF-format remote sensing imagery undergoes content-aware processing: colorful visual-spectrum images are converted to PNG/JPG formats to ensure model compatibility and prevent parsing failures. Conversely, images dominated by non-visible spectral bands are transformed into concatenated grayscale blocks, thereby preserving original spectral information within a grayscale representation.

\item \textbf{Multi-Scale Spatial Cropping:} 
For high-resolution geospatial imagery, we apply randomized cropping based on parcel boundaries~\cite{zhang2026gtpbd}. This strategy generates multi-scale samples that simulate diverse spatial query resolutions, enhancing the model's robustness to scale variations.

\item \textbf{Hierarchical Annotation Extraction:} 
We leverage original labels and masks to derive instance-level statistical metrics and bounding boxes,  creating a standardized dataset featuring hierarchical annotations across three levels: pixel-level (segmentation masks), instance-level (object detection boxes), and parcel-level (agricultural boundaries). To maintain data fidelity, original image resolutions are preserved ranging from 300$\times$300 to 4,500$\times$4,500 pixels.

\item \textbf{Strict Benchmark Decontamination:} 
To ensure rigorous and unbiased evaluation, we enforce a strict physical separation between the instruction-tuning corpus (AgroOmni) and AgroMind~\cite{li2025agromind}. We combine three duplicate signals: byte-exact SHA256 hashing, footprint overlap over image extents, and perceptual matching. The perceptual test flags pairs with a SigLIP cosine similarity above 0.98 and confirms a candidate when the mean absolute difference stays below 2. We calibrate both thresholds on random image pairs and hold the false-positive rate under $10^{-3}$. We also group samples by parcel, acquisition sequence, and geolocation, so that one field or one flight path cannot span the training and evaluation splits. These checks yield zero image-level overlap, so the reported performance reflects cross-scale generalization rather than image memorization.

\end{itemize}

\subsection{Question Generation}
\label{suppl::Question Generation}
We propose a Dual-Track QA Generation method to transform multi-level annotations into high-fidelity instruction-tuning data. Leveraging the expert-level information embedded in the raw data, this method employs two complementary strategies—“Rule-based Generation” and “Human-based Logic Synthesis”—to automatically generate QA pairs covering the full cognitive spectrum from basic perception to complex reasoning.

\textbf{Rule-based Generation.}
For tasks demanding rigorous numerical and positional accuracy, such as Spatial Perception and Object Understanding tasks, we employ a deterministic template filling strategy. Utilizing JSON metadata or segmentation masks extracted during pre-processing (e.g., precise polygon coordinates, CRS projections, and geolocation), we construct a library of parameterized templates. For instance, boundary detection questions are generated by directly formatting normalized coordinates, while climate zone reasoning is derived by mapping geographical coordinates to climate labels. This paradigm relies entirely on ground truth, fundamentally eliminating the numerical and spatial hallucinations often exhibited by LMMs.

\textbf{Human-based Logic Synthesis.}
For tasks involving complex agronomic decision-making, such as Scene Understanding and Scene Reasoning tasks, we translate raw annotations into dynamic reasoning processes that simulate human experts, guiding the model through pre-defined logical chains. We adopt a “reverse-engineering” approach from evidence to conclusion: starting with ground-truth annotations (e.g., “sparse and small weed clusters”), we apply agronomic principles to deduce valid conclusions (e.g., “early-stage infestation requiring preventive weeding”) and formulate corresponding decision-making questions. We require the model to replicate this observational path by first identifying specific visual evidence, then deriving the answer through logical deduction. This strategy balances semantic richness with factual accuracy while mitigating the hallucination risks associated with open-ended generation.

\textbf{Annotation protocol and audit.} The taxonomy, the templates, and the reasoning rules were designed by twelve researchers. Their backgrounds cover agriculture, plant protection, remote sensing, geographic information systems, and computer vision. We audited a stratified sample of 30\% of the corpus. Every sampled pair covers one of the 56 task categories.

Two researchers reviewed each pair independently for format validity, image--annotation correspondence, semantic validity, and task correctness. We kept a pair only when both reviewers agreed. Before retention, we filtered 3.32\% of the candidate pairs as duplicates or low-quality records. For deterministic tasks, we recomputed the answer from the source annotation. The annotation guidelines and the audit protocol are released with the dataset.

\subsection{Processing for Each Dataset}
\label{suppl::detail data}
%重新引用table1，逐段介绍13个数据集
For each dataset in Table~\ref{tab:dataset_summary}, we adopt specific processing methods to design tailored question-answer pairs that align with the given scenarios.

\textbf{ACFR Orchard Fruit Dataset.}
To transform the raw CSV-based fruit radius annotations into structured spatial queries, we first derive the equivalent radius for each fruit instance to calibrate for instance density, filtering out anomalous size outliers. To elevate the reasoning complexity, we partition each image into four discrete spatial quadrants, enabling the generation of queries focused on region-level density distributions and relative spatial patterns. Furthermore, we implement a targeted augmentation strategy by superimposing synthetic fruit patches as distractors, specifically designed to challenge the model's discriminative ability in multi-choice species classification tasks.

\textbf{2018 AI Challenge Dataset.} At first we refine the raw leaf imagery by discarding augmented artifacts, such as rotated, flipped, or redundant files flagged by filename metadata. We then map the original categorical labels—encompassing species, disease types, and severity levels—onto spatial disease distributions to instantiate diverse QA templates. These templates support multimodal inference tasks including crop classification, health-status verification, and severity quantification. To mitigate class imbalance, we apply stratified downsampling across all crop-disease combinations, ensuring a uniform representation of disease progression and species diversity in the final balanced corpus.

\textbf{IP102 Dataset. }We restructure the VOC-formatted annotations of this large-scale pest collection to develop a fine-grained recognition and reasoning corpus. To facilitate the learning of nuanced taxonomic features, we organize identification tasks by pairing target pests with visually similar distractor species from the same superclass. This comparative learning strategy encourages the model to focus on subtle morphological distinctions rather than surface-level appearance. Beyond classification, we integrate pest count labels into the instruction-tuning process to train the model's quantitative reasoning abilities. By linking specific pest identities with their associated host crops, we synthesize conversational pairs that not only evaluate semantic recognition but also enhance the model's foundational knowledge regarding pest-crop ecological relationships.

\textbf{Agriculture Vision Challenge. }We utilize the original six-category binary masks—encompassing cloud shadows, planting irregularities, water-related anomalies, and weed clusters—alongside their corresponding multi-spectral (RGB+NIR) image pairs. To transform these pixel-level labels into rich instructional signals, we derive secondary localization annotations through contour extraction and quantified metrics for regional area and spatial distribution. These derived statistics, when combined with the original boundary masks, serve as the quantitative foundation for instruction-tuning queries focused on anomaly detection and coverage estimation. Moreover, we identify scenes featuring multiple co-occurring anomalies and apply relational positional annotations to evaluate relative spatial arrangements. This process results in a comprehensive training corpus that compels the model to internalize the interplay between multi-spectral visual features and complex spatial configurations in precision agriculture.

\textbf{OilPalmUAV Dataset. }
Initially, we process the original bounding-box annotations by randomly scaling them by a factor of 1.1–1.5 using the PIL ImageDraw module. This augmentation step effectively prevents the model from relying on trivial box-based cues for recognition. Leveraging these refined instances, we curate fine-grained growth-stage classification tasks for individual trees and perform exhaustive instance enumeration to derive scene-level density metrics for quantitative reasoning.

\textbf{CropHarvest Dataset.}
To process the multi-spectral spatiotemporal sequences~\cite{zhang2026gtpbdmm}, we convert the raw GeoTIFF data into normalized band mosaics. This pipeline involves parsing Sentinel-1/2 inputs via GDAL, applying per-band min-max normalization to mitigate sensor-induced variations, and filtering out degenerate observations (e.g., uniform-value bands or invalid pixels) to maintain high data fidelity. Utilizing the resultant spectral mosaics and polygon geometries, we formulate two distinct instruction-tuning modalities: a balanced binary classification task for agricultural land-use verification (comprising 50\% crop and 50\% non-crop samples) and a spatial quantification task based on polygon-counting to assess field fragmentation patterns. These tasks are designed to train the model in interpreting both spectral characteristics and complex land-use layouts at a global scale.

\textbf{GWHD Dataset.}
We unify the dataset formats and map each image to its corresponding metadata, including geographical information and wheat growth stages. By parsing the provided bounding-box annotations, we isolate individual wheat heads to facilitate precise object counting. We subsequently construct four distinct categories of rule-based QA pairs: single-image multiple-choice questions regarding growth stages; dual-image yes/no judgments; multi-image growth stage selections; and cross-image quantity difference calculations, prioritizing pairs with a count variance of no more than 20. These QA pairs are organized with unique identifiers and hierarchical labels, providing a rigorous training framework for evaluating object counting, scene understanding, and spatial reasoning in wheat-centric agricultural environments.

\textbf{CLCD Dataset. } To address the necessity of dynamic agricultural monitoring, we utilize the provided bi-temporal semantic masks to construct high-order temporal reasoning benchmarks. By performing image registration and differential retrieval on the same land parcels across distinct years, we capture granular land-use transitions. Specifically, we calculate the density of changed pixels relative to the global parcel scale to generate reasoning-intensive QA pairs regarding land-use evolution trajectories and change dynamics. This process effectively upgrades the model's perception from static land-cover observation to dynamic evolutionary sensing, which is essential for mastering the long-term temporal characteristics of agricultural landscapes.

\textbf{PhenoBench Dataset.}
In this dataset, we process the semantic segmentation masks to explicitly distinguish between intact and partially visible crop or weed instances. By aggregating instance-level annotations, we derive precise counts for individual plants and leaves. Furthermore, we utilize pixel-level visibility maps to quantify plant completeness and inter-plant occlusion, establishing a robust quantitative foundation for tasks spanning plant counting, occlusion reasoning, and competitive weed-crop interaction analysis. This structured conversion enables the model to learn complex spatial and occlusion-aware concepts essential for precision agricultural monitoring.

\textbf{Tomato Dataset. }We parse dual-format annotations (JSON and VOC+XML) across four distinct lighting conditions—natural, artificial, faint, and sodium-yellow laser—to extract instance-level bounding boxes and three-tier maturity labels (unripe, half-ripe, ripe).  By utilizing the extracted maturity metadata, we compute the ratio of fully ripe tomatoes per scene to instantiate comparative visual QA pairs, assessing which of two randomly sampled images exhibits a higher overall crop maturity. Additionally, we aggregate instance-level labels to formulate multiple-choice queries identifying the exact subset of ripening stages present within a given frame. Finally, to evaluate global spatial reasoning, we calculate a single maximum-encompassing bounding box by aggregating the extreme coordinates of all individual instances, generating localization queries that require the model to encapsulate the entire tomato cluster simultaneously.

\textbf{EarthVQA Dataset. }To ground the model's spatial cognition in complex agricultural environments, we leverage the eight provided semantic category labels. Beyond conventional classification, we perform geometric analysis on these pixel-level masks by computing the geometric centroids of individual land parcels. By comparing these spatial landmarks, we construct logical reasoning QA pairs that require the model to interpret relative spatial configurations and layouts. This transformation shifts the training focus from mere "entity recognition" to "spatial topology understanding," effectively enhancing the model's capacity for relational geometric reasoning.

\textbf{OAM-TCD Dataset.}
In this dataset, we curate a training subset converting raw GeoTIFF tree-cover maps into RGB-compressed JPEGs to optimize visual fidelity for multimodal encoders. Leveraging the original semantic annotations, we establish foundational tasks spanning ecological region classification and individual plant identification. To synthesize complex spatial reasoning, we implement a 3×3 grid-based partitioning scheme, computing per-cell canopy density to derive regional distribution metrics. These spatial statistics enable advanced queries concerning global tree-cover patterns, while the random sampling of multi-grid cell pairs allows for the construction of multi-image comparison tasks that train the model to reason across heterogeneous landscape layouts.

\section{Model Details of AgroMind Evaluation}
\label{suppl:agromind_eval_details}

To establish a rigorous and comprehensive evaluation baseline, we curate a diverse suite of MLLMs representing distinct paradigms. This structured taxonomy systematically isolates the performance delta between generic pre-training, existing domain patches, and our AgroOmni-driven alignment strategy. The evaluated models are detailed as follows:

\textbf{LLaVA-NeXT}~\cite{llavanext} employs a dynamic high-resolution image processing strategy and an upgraded AnyRes technique to preserve detailed visual semantics. It is included as a representative general-purpose architecture from the original AgroMind benchmark to evaluate standard zero-shot spatial reasoning capabilities and multi-patch visual comprehension on complex agricultural scenes.

\textbf{InternVL-2}~\cite{internvl2} scales up the vision foundation model with a progressive alignment strategy and supports dynamic resolution scaling. Featuring a highly capable LLM backbone, it serves as a leading general-purpose foundational model to establish a robust and highly competitive benchmark performance for the current open-source ecosystem.

\textbf{GeoChat}~\cite{geochat} is a specialized foundational VLM explicitly adapted for remote sensing applications. Pre-trained on a massive corpus of high-resolution multi-sensor imagery, it is included to test the limits and potential overfitting of top-down orthophoto optimization when confronted with the heterogeneous, multi-view (Ground, UAV, and Satellite) agricultural scenarios present in our benchmark.

\textbf{GPT-5.2}~\cite{gpt-5} represents the latest generation of state-of-the-art proprietary models, featuring massive-scale multimodal pre-training and unparalleled general world knowledge. It is evaluated to probe the absolute zero-shot upper bound for complex visual reasoning, geometric logic, and cross-scale spatial grounding in precision agriculture.

\textbf{Claude-4.6-Sonnet}~\cite{claudesonnet46} is incorporated alongside GPT-5.2 to comprehensively evaluate the performance ceiling of industry-leading, closed-source systems. It provides a critical comparative benchmark for advanced visual parsing, nuanced instruction-following, and reasoning over dense visual environments.

\textbf{AgriGPT-VL}~\cite{agrigpt_vl} serves as a domain-specific reference model tailored for agronomic tasks. It is utilized to contextualize prior efforts in multimodal agricultural adaptation and to highlight the inherent limitations of models trained predominantly on conventional, single-perspective (terrestrial-centric) agricultural data.

\textbf{AgroOmni Baselines (NVILA)}~\cite{nvila} are established by applying our complete alignment pipeline—spanning supervised fine-tuning (SFT) and Group Relative Policy Optimization (GRPO)—to the base NVILA-Lite-8B architecture. Equipped with a frozen SigLIP vision encoder, these baselines are trained exclusively on the AgroOmni corpus to definitively validate the efficacy, high-quality reward signals, and cross-scale reasoning potential of our multi-view dataset.

\section{Qualitative Analysis and Case Studies}
\label{suppl:case_study}

\begin{figure}[t]
    \centering
    % Case 1
    \includegraphics[width=\textwidth, height=0.42\textheight, keepaspectratio]{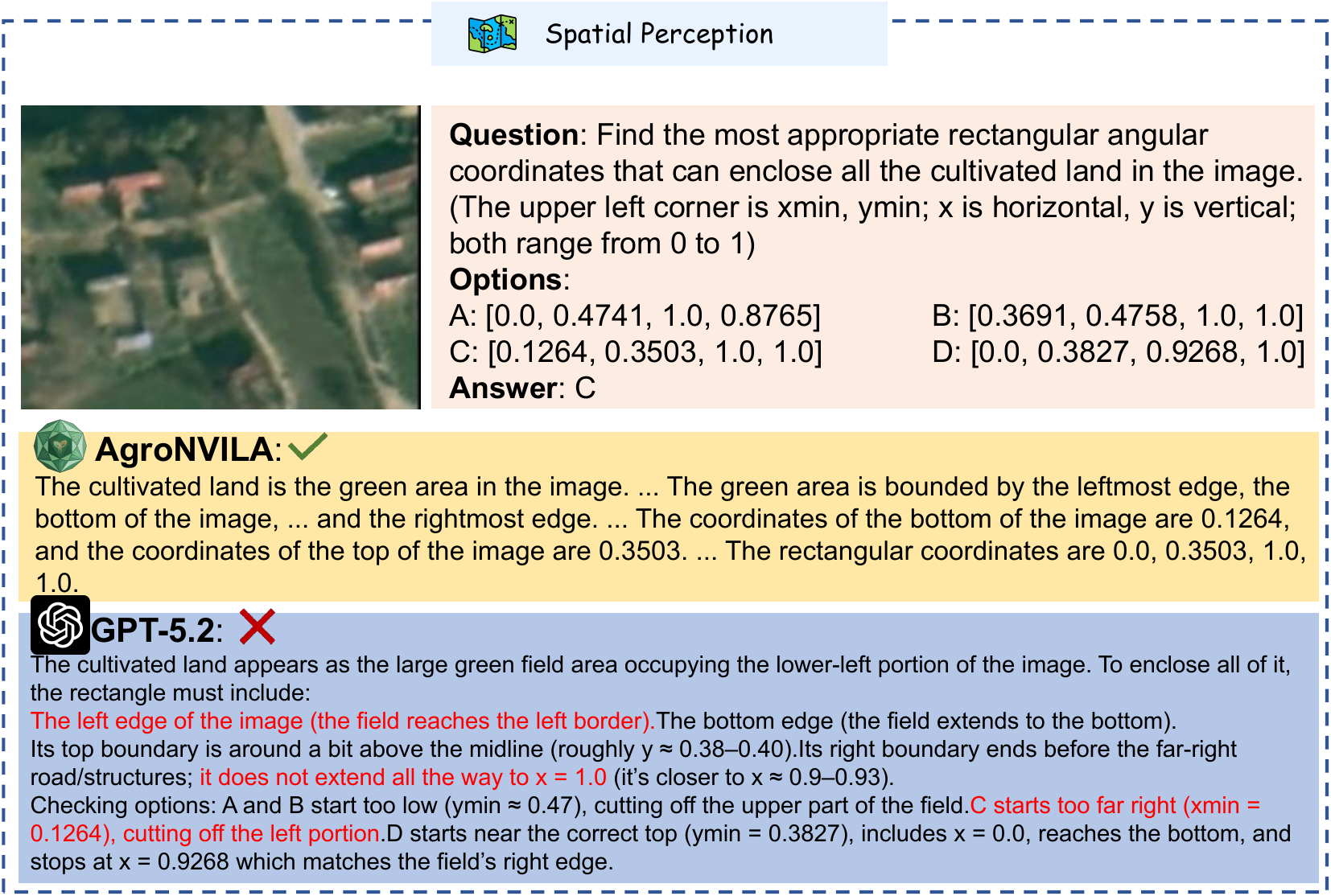}
    \vspace{-0.5cm}
    \caption{\textbf{Qualitative Result on Spatial Perception.} When enclosing cultivated land, the baseline model (GPT-5.2) completely inverts the spatial mapping, incorrectly anchoring the target to the lower-left corner. AgroNVILA accurately grounds the semantic concept to the correct normalized spatial coordinates.}
    \label{fig:case_spatial}
    
    \vspace{0.5cm} % 强行拉开上下两张图的呼吸间距
    
    % Case 2
    \includegraphics[width=\textwidth, height=0.42\textheight, keepaspectratio]{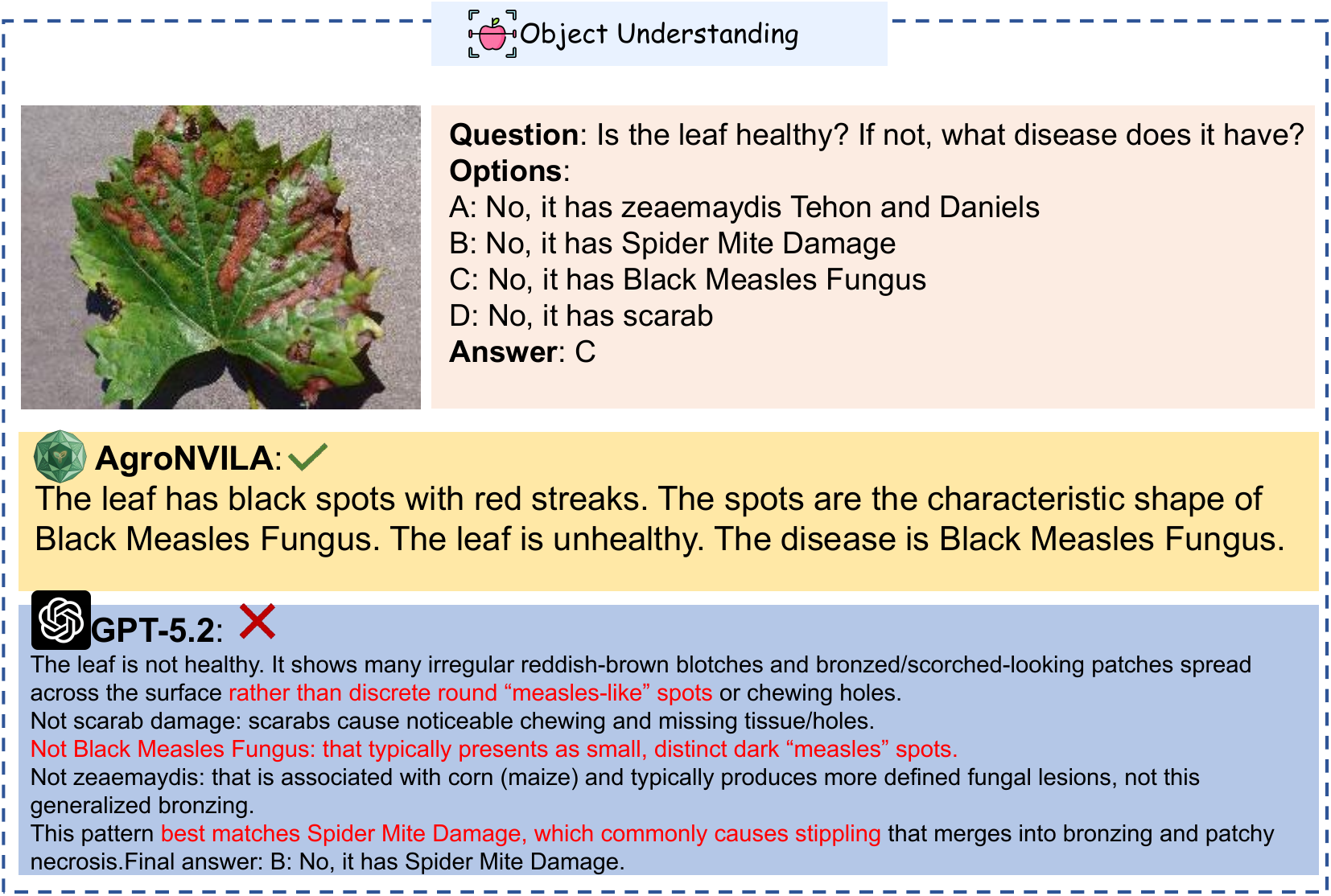}
    \vspace{-0.5cm}
    \caption{\textbf{Qualitative Result on Object Understanding.} In fine-grained pathological diagnosis, the baseline relies on generic visual priors, misinterpreting the symptoms as Spider Mite Damage. AgroNVILA successfully captures high-frequency pathological details (e.g., black spots and red streaks) to correctly diagnose Black Measles Fungus.}
    \label{fig:case_object}
\end{figure}

\begin{figure}[t]
    \centering
    % Case 3
    \includegraphics[width=\textwidth, height=0.42\textheight, keepaspectratio]{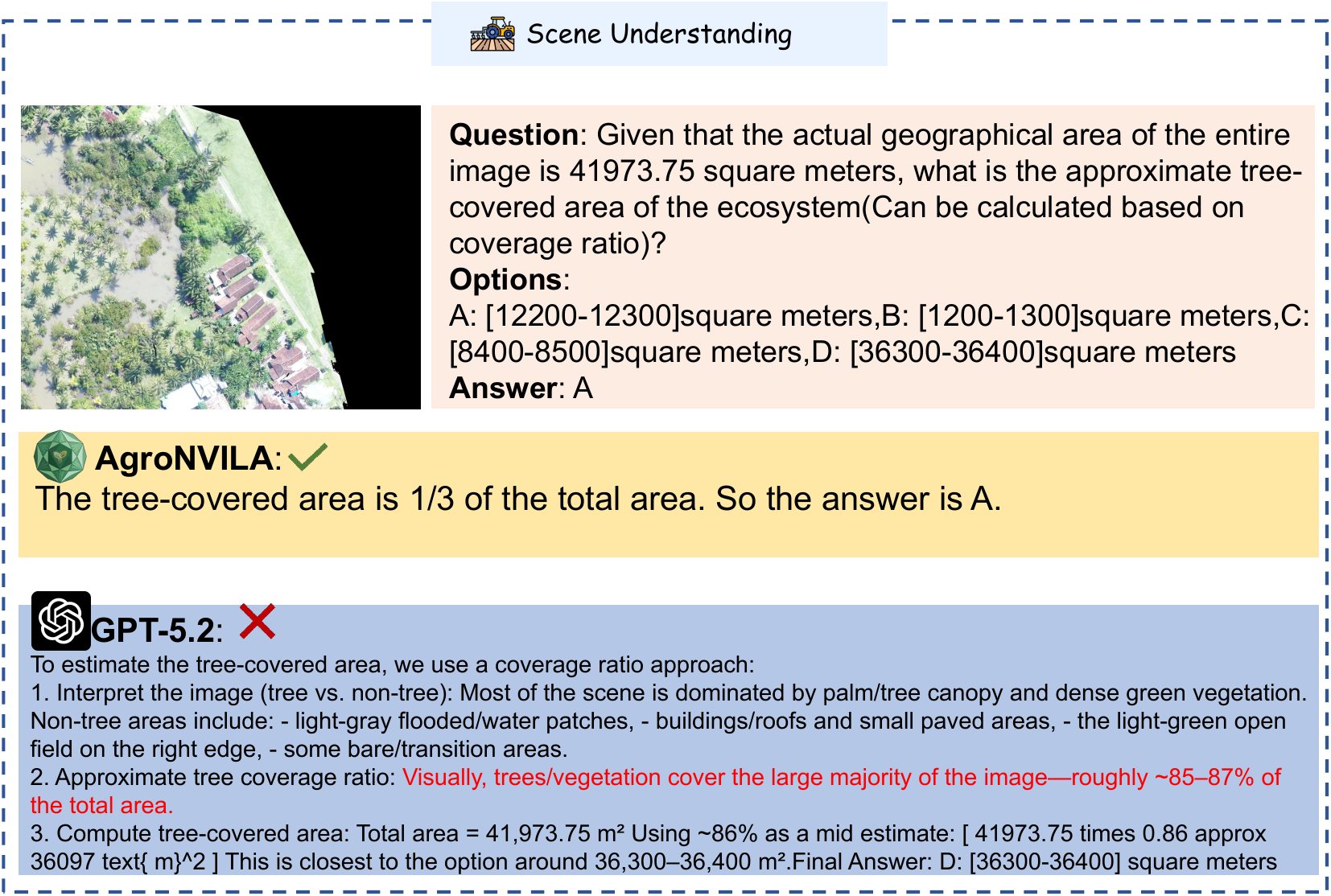}
    \vspace{-0.5cm}
    \caption{\textbf{Qualitative Result on Scene Understanding.} For macro-area coverage estimation, the baseline model suffers from severe visual illusions, drastically overestimating the tree coverage ratio to $\sim$86\%. AgroNVILA bypasses verbose hallucinations and accurately outputs the calibrated target area of 12,200-12,300 square meters.}
    \label{fig:case_scene_understanding}
    
    \vspace{0.5cm} % 强行拉开上下两张图的呼吸间距
    
    % Case 4
    \includegraphics[width=\textwidth, height=0.42\textheight, keepaspectratio]{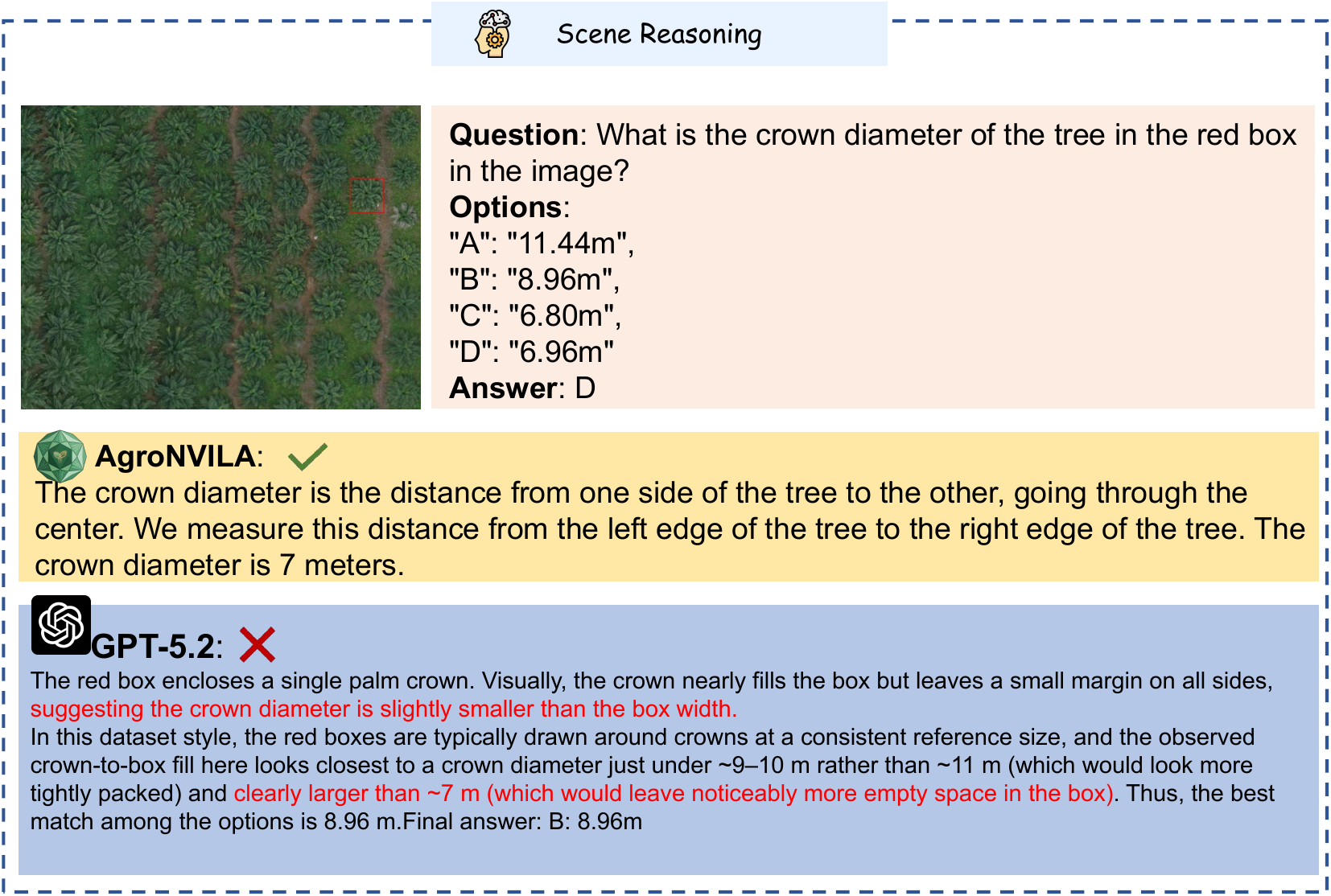}
    \vspace{-0.5cm}
    \caption{\textbf{Qualitative Result on Scene Reasoning.} When measuring absolute crown diameter, the baseline lacks a physical scale prior and resorts to blind guessing based on bounding box margins. AgroNVILA demonstrates strong implicit scale awareness, successfully anchoring the visual representation to the precise metric scale of 6.96m.}
    \label{fig:case_scene_reasoning}
\end{figure}

While quantitative metrics provide a macro-level evaluation of model performance, they often obscure the underlying reasoning mechanisms and visual grounding capabilities of Multimodal Large Language Models (MLLMs). To investigate the domain gap between general-purpose vision-language models and our domain-specific AgroNVILA, we conduct a detailed qualitative analysis using chain-of-thought prompting. As illustrated in Fig.~\ref{fig:case_spatial}, \ref{fig:case_object}, \ref{fig:case_scene_understanding} and \ref{fig:case_scene_reasoning}, we contrast the critical reasoning errors made by the second-best model with the precise visual grounding achieved by AgroNVILA across four fundamental agricultural perception tasks.

\paragraph{Case 1: Spatial Coordinate Alignment.} 
In the bounding box regression task (enclosing cultivated land), the second-best model demonstrates a complete inversion of spatial mapping, incorrectly anchoring the semantic concept of "cultivated land" to the lower-left corner. Conversely, AgroNVILA exhibits precise semantic-to-coordinate mapping. It accurately defines the visual attributes ("green and has no buildings") and seamlessly translates this semantic understanding into the correct normalized spatial coordinates $(0.1264, 0.3503)$ to $(1.0, 1.0)$.

\paragraph{Case 2: Domain-Specific Pathological Diagnosis.} 
When identifying fine-grained leaf diseases, the second-best model attempts to map visual symptoms using general-domain common sense, misinterpreting the irregular blotches as "Spider Mite Damage." AgroNVILA, however, demonstrates profound pathological feature extraction. It accurately captures the critical high-frequency details—"black spots and red streaks"—and explicitly links these symptoms to the etiology of "Black Measles Fungus," proving the effectiveness of our domain-specific feature alignment.

\paragraph{Case 3: Global Proportion Estimation.} 
For the macro-area coverage estimation task, the second-best model exhibits severe visual illusions, drastically overestimating the tree coverage ratio (guessing $\sim$86\% instead of the ground truth $\sim$29\%). In stark contrast, AgroNVILA successfully overcomes this integration bottleneck. While its textual output is highly concise—a common trait of models optimized for strict QA correctness over conversational verbosity—its visual estimation is perfectly calibrated, accurately identifying the target range of 12,200-12,300 square meters.

\paragraph{Case 4: Absolute Physical Scale Anchoring.} 
In the crown diameter measurement task, the second-best model attempts to infer absolute physical dimensions purely based on the relative empty space within the bounding box, resulting in blind guessing. AgroNVILA successfully anchors the visual representation to the correct physical metric scale (6.96m). This indicates that our pre-training and adaptation strategy effectively endows the model with implicit scale awareness, a capability fundamentally lacking in generic MLLMs.

\vspace{0.5cm}
\noindent \textbf{Conclusion.} 
These comparative cases empirically validate that the second-best's performance degradation is a systematic breakdown in spatial grounding, scale perception, and pathological priors. AgroNVILA effectively recalibrates these representations. Furthermore, our model significantly suppresses the conversational hallucinations typical of generic LLMs, trading verbosity for rigorous, domain-accurate decision-making.

% --- 第二页 Case Study (包含 Case 3 和 Case 4) ---

\section{Broader Impacts}
\label{suppl:bi}
The deployment of multi-view agricultural multimodal models, such as AgroNVILA, carries profound societal implications. On a positive note, democratizing expert-level agronomic reasoning can significantly enhance global food security and economic stability by providing scalable, precise interventions to farmers worldwide. However, the deployment of such models must be managed carefully to mitigate potential negative impacts. Over-reliance on automated systems could lead to the degradation of traditional, localized agronomic knowledge. Furthermore, while the models are designed for environmental monitoring and yield optimization, macro-scale remote sensing capabilities could theoretically be misappropriated for unauthorized surveillance or competitive intelligence in global agricultural markets. Ensuring equitable access to these technologies is critical to prevent widening the technological gap between large-scale agribusinesses and smallholder farmers.

\section{Limitations and Future Work}
\label{suppl:laf}
While AgroOmni provides a high-density learning signal, our diagnostic evaluations identify a persistent structural lag in Spatial Perception tasks, revealing an inherent architectural ceiling in current multimodal frameworks. Supervised fine-tuning effectively aligns semantic reasoning but struggles to completely overcome the fundamental absence of macro-level spatial priors in frozen generic vision encoders. Future research must prioritize spatial-centric architectural alignment to better process top-down topologies. Subsequent work will also focus on enhancing model scalability and deploying AgroNVILA within real-world, dynamic agricultural systems to validate its temporal reasoning efficacy in actual production environments.

AgroOmni covers only the regions and crops listed in Table~\ref{tab:coverage}, and we make no claim beyond that coverage. Deployment in other regions, crops, or seasons needs additional validation. Geospatial privacy is discussed in Appendix~\ref{suppl:ec}.

Most of our QA pairs follow parameterized templates, so a model can learn answer priors instead of visual evidence. A template shortcut can inflate scores on tasks with a narrow answer set. The open-ended description in Fig.~\ref{fig:ablationcase} goes beyond the template answer space, but we do not quantify how often this happens. Future work will add free-form expert questions to test it.

\section{Ethical Considerations}
\label{suppl:ec}
This research follows the ethical guidelines for machine learning studies. AgroOmni aggregates public agricultural datasets and a proprietary parcel collection. The collection contains no human subjects, no personally identifiable information, and no crowdsourced labor.

Data ownership and release follow layered licensing. The question--answer annotations and instruction templates that we contribute are released under CC BY-SA 4.0. The imagery keeps the license of its source, and we hold no copyright over it. Content from EarthVQA (CC BY-NC-SA 4.0) and IP102 (academic use only) is not for commercial use. Table~\ref{tab:coverage} lists the sources. The proprietary parcel component is published with the permission of the data owner.

We release imagery and question--answer records only. We distribute no coordinates, no parcel boundaries, and no georeferencing metadata, which limits the risk of re-identification. All datasets and baselines are released for academic research. Training large multimodal models consumes energy, and our use of LoRA and GRPO limits this cost.
% \section{Technical appendices and supplementary material}
% Technical appendices with additional results, figures, graphs, and proofs may be submitted with the paper submission before the full submission deadline (see above). You can upload a ZIP file for videos or code, but do not upload a separate PDF file for the appendix. There is no page limit for the technical appendices. 

% Note: Think of the appendix as ``optional reading'' for reviewers. The paper must be able to stand alone without the appendix; for example, adding critical experiments that support the main claims to an appendix is inappropriate. 

%%%%%%%%%%%%%%%%%%%%%%%%%%%%%%%%%%%%%%%%%%%%%%%%%%%%%%%%%%%%

\newpage
\clearpage
\section*{NeurIPS Paper Checklist}

\begin{enumerate}

\item {\bf Claims}
    \item[] Question: Do the main claims made in the abstract and introduction accurately reflect the paper's contributions and scope?
    \item[] Answer: \answerYes{} % Replace by \answerYes{}, \answerNo{}, or \answerNA{}.
    \item[] Justification: The abstract and introduction clearly state the claims made, including the curation of the multi-view AgroOmni dataset to address ground-level bias and the proposal of the AgroNVILA model to achieve cross-scale agricultural reasoning.

\item {\bf Limitations}
    \item[] Question: Does the paper discuss the limitations of the work performed by the authors?
    \item[] Answer: \answerYes{} % Replace by \answerYes{}, \answerNo{}, or \answerNA{}.
    \item[] Justification: Appendix I (Limitations and Future Work) and Section 5.3 discuss the limitations of this work.

\item {\bf Theory assumptions and proofs}
    \item[] Question: For each theoretical result, does the paper provide the full set of assumptions and a complete (and correct) proof?
    \item[] Answer: \answerNA{} % Replace by \answerYes{}, \answerNo{}, or \answerNA{}.
    \item[] Justification: The paper focuses on dataset curation and empirical multimodal model evaluation; it does not introduce new theoretical results or mathematical proofs.

    \item {\bf Experimental result reproducibility}
    \item[] Question: Does the paper fully disclose all the information needed to reproduce the main experimental results of the paper to the extent that it affects the main claims and/or conclusions of the paper (regardless of whether the code and data are provided or not)?
    \item[] Answer: \answerYes{} % Replace by \answerYes{}, \answerNo{}, or \answerNA{}.
    \item[] Justification: The paper fully details the experimental setup in Section 5.1, provides extensive hyperparameters in Appendix B, and shares a public repository link containing all training scripts for reproducing the results.

\item {\bf Open access to data and code}
    \item[] Question: Does the paper provide open access to the data and code, with sufficient instructions to faithfully reproduce the main experimental results, as described in supplemental material?
    \item[] Answer: \answerYes{} % Replace by \answerYes{}, \answerNo{}, or \answerNA{}.
    \item[] Justification: We provide a public GitHub repository link for the training scripts, and Appendix E explicitly details the 12 public datasets utilized along with respective access links.

\item {\bf Experimental setting/details}
    \item[] Question: Does the paper specify all the training and test details (e.g., data splits, hyperparameters, how they were chosen, type of optimizer) necessary to understand the results?
    \item[] Answer: \answerYes{} % Replace by \answerYes{}, \answerNo{}, or \answerNA{}.
    \item[] Justification: Detailed experimental settings, including the stage-wise training pipeline, optimizer configurations (AdamW), learning rates, and LoRA configurations, are explicitly specified in Appendix B and summarized in Table 6.

\item {\bf Experiment statistical significance}
    \item[] Question: Does the paper report error bars suitably and correctly defined or other appropriate information about the statistical significance of the experiments?
    \item[] Answer: \answerYes{} % Replace by \answerYes{}, \answerNo{}, or \answerNA{}.
    \item[] Justification: Table 9 reports the mean and the standard deviation over three random seeds for the full-corpus SFT configuration. The other tables report single deterministic runs, which is standard practice for the utilized MLLM evaluation protocols.

\item {\bf Experiments compute resources}
    \item[] Question: For each experiment, does the paper provide sufficient information on the computer resources (type of compute workers, memory, time of execution) needed to reproduce the experiments?
    \item[] Answer: \answerYes{} % Replace by \answerYes{}, \answerNo{}, or \answerNA{}.
    \item[] Justification: The paper explicitly states in Appendix B.1 that experiments and ablations were conducted on a computing cluster equipped with NVIDIA A100 (80GB) GPUs.
    
\item {\bf Code of ethics}
    \item[] Question: Does the research conducted in the paper conform, in every respect, with the NeurIPS Code of Ethics \url{https://neurips.cc/public/EthicsGuidelines}?
    \item[] Answer: \answerYes{} % Replace by \answerYes{}, \answerNo{}, or \answerNA{}.
    \item[] Justification: The research utilizes agricultural imagery and metadata to advance precision farming and multimodal reasoning, which conforms to standard ethical guidelines without requiring special deviations.

\item {\bf Broader impacts}
    \item[] Question: Does the paper discuss both potential positive societal impacts and negative societal impacts of the work performed?
    \item[] Answer: \answerYes{} % Replace by \answerYes{}, \answerNo{}, or \answerNA{}.
    \item[] Justification: The societal impacts are discussed in the first paragraph of the introduction, which elaborates the societal needs, practical bottlenecks, and positive benefits of applying multimodal models to intelligent agriculture, and in Appendix H.
    
\item {\bf Safeguards}
    \item[] Question: Does the paper describe safeguards that have been put in place for responsible release of data or models that have a high risk for misuse (e.g., pre-trained language models, image generators, or scraped datasets)?
    \item[] Answer: \answerYes{} % Replace by \answerYes{}, \answerNo{}, or \answerNA{}.
    \item[] Justification: Appendix J states that this work is intended solely for academic research and exploratory practical applications in standardized agricultural scenarios, establishing clear safeguards against inappropriate misuse.

\item {\bf Licenses for existing assets}
    \item[] Question: Are the creators or original owners of assets (e.g., code, data, models), used in the paper, properly credited and are the license and terms of use explicitly mentioned and properly respected?
    \item[] Answer: \answerYes{} % Replace by \answerYes{}, \answerNo{}, or \answerNA{}.
    \item[] Justification: The original datasets are properly credited with citations, and their direct access URLs and content details are provided in Table 11. 

\item {\bf New assets}
    \item[] Question: Are new assets introduced in the paper well documented and is the documentation provided alongside the assets?
    \item[] Answer: \answerYes{} % Replace by \answerYes{}, \answerNo{}, or \answerNA{}.
    \item[] Justification: The newly introduced AgroOmni dataset is comprehensively documented across Sections 3, 4, and Appendix E, which details data collection, pre-processing, question generation, and multi-view composition.

\item {\bf Crowdsourcing and research with human subjects}
    \item[] Question: For crowdsourcing experiments and research with human subjects, does the paper include the full text of instructions given to participants and screenshots, if applicable, as well as details about compensation (if any)? 
    \item[] Answer: \answerNA{} % Replace by \answerYes{}, \answerNo{}, or \answerNA{}.
    \item[] Justification: The research is entirely based on agricultural multi-sensor imagery (Ground, UAV, Satellite)  and does not involve human subjects or crowdsourcing experiments.

\item {\bf Institutional review board (IRB) approvals or equivalent for research with human subjects}
    \item[] Question: Does the paper describe potential risks incurred by study participants, whether such risks were disclosed to the subjects, and whether Institutional Review Board (IRB) approvals (or an equivalent approval/review based on the requirements of your country or institution) were obtained?
    \item[] Answer: \answerNA{} % Replace by \answerYes{}, \answerNo{}, or \answerNA{}.
    \item[] Justification: The study does not involve human subjects; therefore, IRB approval is inapplicable.

\item {\bf Declaration of LLM usage}
    \item[] Question: Does the paper describe the usage of LLMs if it is an important, original, or non-standard component of the core methods in this research? Note that if the LLM is used only for writing, editing, or formatting purposes and does \emph{not} impact the core methodology, scientific rigor, or originality of the research, declaration is not required.
    %this research? 
    \item[] Answer: \answerNA{} % Replace by \answerYes{}, \answerNo{}, or \answerNA{}.
    \item[] Justification: The core methodology involves training a domain-specific vision-language model using rule-based and human-based logic synthesis; off-the-shelf proprietary LLMs are solely utilized as standard evaluation baselines  and not as undocumented core generative components.

\end{enumerate}

\end{document}